\documentclass[11pt, a4paper]{deepmind}

\title{Learning Agile Soccer Skills for a Bipedal Robot with Deep Reinforcement Learning\footnote{This is the author's version of the work. It is posted here by permission of the AAAS for personal use, not for redistribution. The definitive version \url{https://www.science.org/doi/10.1126/scirobotics.adi8022} was published in Science Robotics on 2024-04-10, DOI: 10.1126/scirobotics.adi8022.}}

\correspondingauthor{tuomash@google.com, guylever@google.com}

\reportnumber{} % Leave blank if n/a

\author{
Tuomas Haarnoja,$^{1*+}$
Ben Moran,$^{1*}$
Guy Lever,$^{1*+}$
Sandy H. Huang,$^{1*}$
Dhruva Tirumala,$^{12}$
Jan Humplik,$^{1}$
Markus Wulfmeier,$^{1}$
Saran Tunyasuvunakool,$^{1}$
Noah Y. Siegel,$^{1}$
Roland Hafner,$^{1}$
Michael Bloesch,$^{1}$
Kristian Hartikainen,$^{3}$
Arunkumar Byravan,$^{1}$
Leonard Hasenclever,$^{1}$
Yuval Tassa,$^{1}$
Fereshteh Sadeghi,$^{4}$
Nathan Batchelor,$^{1}$
Federico Casarini,$^{1}$
Stefano Saliceti,$^{1}$
Charles Game,$^{1}$
Neil Sreendra,$^{}$
Kushal Patel,$^{}$
Marlon Gwira,$^{}$
Andrea Huber,$^{1}$
Nicole Hurley,$^{5}$
Francesco Nori,$^{1}$
Raia Hadsell,$^{1}$
Nicolas Heess,$^{1}$
\\
$^1$ Google DeepMind\\
$^2$ University College London\\
$^3$ Work completed at DeepMind; current affiliation: University of Oxford\\
$^4$ Work completed at DeepMind; current affiliation: Google\\
$^5$ Work completed at DeepMind; current affiliation: Isomorphic Labs\\
$^*$ Equal contribution\\
$^+$To whom correspondence should be addressed;\\
E-mail: tuomash@google.com, guylever@google.com
}

\usepackage{multirow} % for tables
\usepackage{booktabs} % for tables
\usepackage{wrapfig}

\newcommand{\SI}[2]{#1\,#2}
\newcommand{\percent}{\%}
\newcommand{\m}{m}
\newcommand{\meter}{m}
\newcommand{\s}{s}
\newcommand{\radian}{rad}
\newcommand{\ms}{ms}
\newcommand{\hertz}{Hz}
\newcommand{\cm}{cm}
\newcommand{\kg}{kg}
\newcommand{\per}{/}
\newcommand{\N}{N}
\newcommand{\squared}{$^2$}
\newcommand{\degree}{$^\circ$}
\newcommand{\mm}{mm}

\usepackage[font=footnotesize]{caption}
\usepackage{xcolor}
\newif\ifshowcomments
\usepackage{subcaption}
\usepackage{units}
\usepackage{graphicx}
\usepackage{hyperref}
\usepackage{amssymb}
\usepackage{amsmath}
\usepackage{bbm}
\usepackage{float}

\newcommand{\cA}{\mathcal{A}}
\newcommand{\cS}{\mathcal{S}}
\newcommand{\cJ}{\mathcal{J}}

\newcommand{\clambdas}{\lambda_\mathrm{s}}
\newcommand{\clambdag}{\lambda_\mathrm{g}}
\newcommand{\cQs}{Q_\mathrm{s}}
\newcommand{\cQg}{Q_\mathrm{g}}
\newcommand{\cpis}{\pi_\mathrm{s}}
\newcommand{\cpig}{\pi_\mathrm{g}}
\newcommand{\cU}{\mathcal{U}}
\newcommand{\traj}{\xi}

\newcommand{\ot}{\boldsymbol{o}{_t}}
\newcommand{\st}{\boldsymbol{s}{_t}}
\newcommand{\stp}{\boldsymbol{s}{_{t+1}}}
\newcommand{\sz}{\boldsymbol{s}{_0}}
\newcommand{\at}{\boldsymbol{a}{_t}}
\newcommand{\ut}{\boldsymbol{u}{_t}}
\newcommand{\utm}{\boldsymbol{u}{_{t-1}}}

\DeclareMathOperator*{\argmax}{argmax}
\DeclareMathOperator*{\argmin}{argmin}

\renewcommand{\eqref}[1]{\hyperref[#1]{Equation~\ref*{#1}}}
\newcommand{\sref}[1]{\hyperref[#1]{Section~\ref*{#1}}}
\newcommand{\aref}[1]{\hyperref[#1]{Suppl. \nameref*{#1}}}
\newcommand{\figureref}[1]{\hyperref[#1]{Figure~\ref*{#1}}}
\newcommand{\supfigureref}[1]{\hyperref[#1]{Suppl.~Figure~\ref*{#1}}}
\newcommand{\algoref}[1]{\hyperref[#1]{Algorithm~\ref*{#1}}}
\newcommand{\tableref}[1]{\hyperref[#1]{Table~\ref*{#1}}}
\newcommand{\suptableref}[1]{\hyperref[#1]{Suppl.~Table~\ref*{#1}}}
\newcommand{\extendedFigName}{{Suppl. Figure}}
\newcommand{\extendedTableName}{{Suppl. Table}}
\newcommand{\movName}{{Movie}}
\newcommand{\supMovName}{{Suppl. Movie}}
\newcommand{\tabref}[1]{\extendedTableName~\ref{#1}}
\newcommand{\xfigref}[1]{\extendedFigName~\ref{#1}}
\newcommand{\supmovref}[1]{\supMovName~#1}
\newcommand{\movref}[1]{\movName~#1}

\newcommand{\citep}{\cite}
\newcommand{\citet}{\cite}

\newcommand{\movoverview}{1}
\newcommand{\movbehaviors}{2}
\newcommand{\movbaselinecomparison}{3}
\newcommand{\movslomoturning}{4}
\newcommand{\movtraininginsim}{S1} 
\newcommand{\movmatches}{S2}
\newcommand{\movsetpieces}{S3}
\newcommand{\movpushes}{S4}
\newcommand{\movvision}{S5}
\newcommand{\beginsupplement}{
  \setcounter{section}{0}
  \renewcommand{\thesection}{S\arabic{section}}
  \setcounter{table}{0}  
  \renewcommand{\thetable}{S\arabic{table}}
  \setcounter{figure}{0} 
  \renewcommand{\thefigure}{S\arabic{figure}}
}

\usepackage{times}

\makeatletter
\def\ttl@useclass#1#2{%
  \@ifstar
    {\ttl@labelfalse\@dblarg{#1{#2}}}
    {\ttl@labeltrue\@dblarg{#1{#2}}}}
\makeatother

\begin{abstract}
We investigate whether Deep Reinforcement Learning (Deep RL) is able to synthesize sophisticated and safe movement skills for a low-cost, miniature humanoid robot that can be composed into complex behavioral strategies in dynamic environments.  We used Deep RL to train a humanoid robot with 20 actuated joints to play a simplified one-versus-one (1v1) soccer game. The resulting agent exhibits robust and dynamic movement skills such as rapid fall recovery, walking, turning, kicking and more; and it transitions between them in a smooth, stable, and efficient manner. The agent's locomotion and tactical behavior adapts to specific game contexts in a way that would be impractical to manually design. The agent also developed a basic strategic understanding of the game, and learned, for instance, to anticipate ball movements and to block opponent shots. Our agent was trained in simulation and transferred to real robots zero-shot. We found that a combination of sufficiently high-frequency control, targeted dynamics randomization, and perturbations during training in simulation enabled good-quality transfer. Although the robots are inherently fragile, basic regularization of the behavior during training led the robots to learn safe and effective movements while still performing in a dynamic and agile way---well beyond what is intuitively expected from the robot. Indeed, in experiments, they walked \SI{181}{\percent} faster, turned \SI{302}{\percent} faster, took \SI{63}{\percent} less time to get up, and kicked a ball \SI{34}{\percent} faster than a scripted baseline, while efficiently combining the skills to achieve the longer term objectives.
\end{abstract}

\begin{document}

\maketitle

% \tableofcontents

\section*{Introduction}
Creating general embodied intelligence, that is, creating agents that can act in the physical world with agility, dexterity, and understanding---as animals or humans do---is one of the long-standing goals of AI researchers and roboticists alike. Animals and humans are not just masters of their bodies, able to perform and combine complex movements fluently and effortlessly, but they also perceive and understand their environment and use their bodies to effect complex outcomes in the world.

Attempts at creating intelligent embodied agents with sophisticated motor capabilities go back many years, both in simulation \citep{sims} and in the real world \citep{raibert1986legged, kuindersma2016optimization}. Progress has recently accelerated considerably, and learning-based approaches have contributed substantially to this acceleration \citep{peters2008reinforcement,deisenroth2013survey,kober2013reinforcement}. In particular, deep reinforcement learning (deep RL) has proven capable of solving complex motor control problems for both simulated characters \citep{heess2017emergence,bansal2017emergent,peng2018deepmimic,merel2020,LiuHumanoidSoccer} and physical robots. High-quality quadrupedal legged robots have become widely available and have been used to demonstrate behaviors, ranging from robust~\cite{lee2020learning,choi2023learning} and agile~\cite{hwangbo2019learning,peng2020learning} locomotion to fall recovery~\cite{lee2019robust}, climbing~\cite{rudin2022advanced}, and basic soccer skills such as dribbling~\cite{ji2023dribblebot,bohez2022imitate}, shooting~\cite{ji2022hierarchical}, intercepting~\cite{huang2022creating} or catching~\cite{forrai2023event} a ball, and simple manipulation with legs~\cite{cheng2023legs}. On the other hand, much less work has been dedicated to the control of humanoids and bipeds, which impose additional challenges around stability, robot safety, number of degrees of freedom, and availability of suitable hardware. The existing learning-based work has been more limited and focused on learning and transfer of distinct basic skills such as walking~\cite{xie2019iterative}, running~\cite{agility2022cassie}, stair climbing~\cite{siekmann2021blind}, and jumping~\cite{LiBipedalJumping}. The state-of-the-art in humanoid control uses targeted model-based predictive control~\cite{deits2023picking}, thus limiting the generality of the method. 

\begin{figure}[H]
  \centering
   \includegraphics[width=\textwidth]{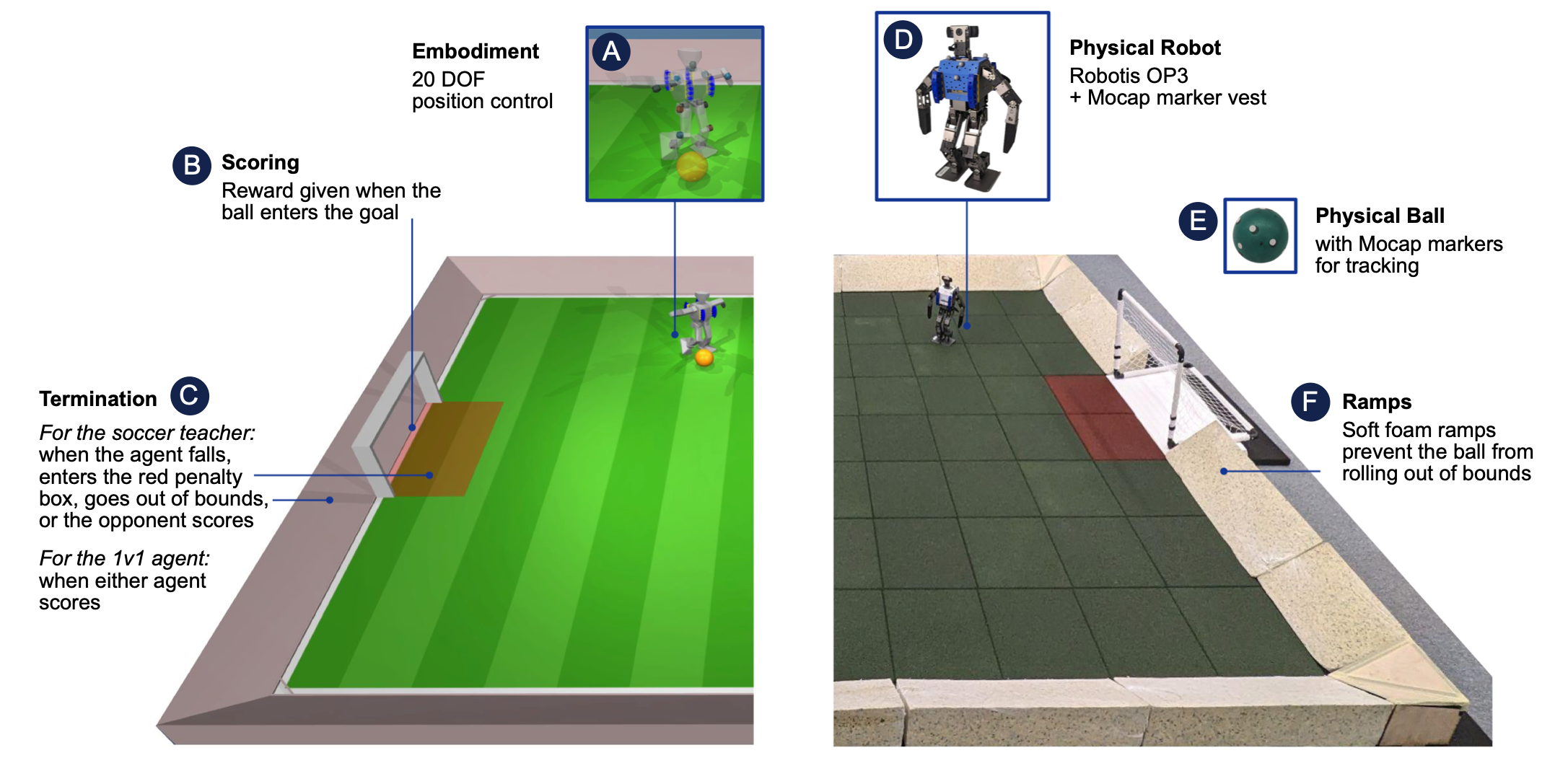}
  \caption{\textbf{The robot soccer environment.} We created matching simulated (left) and real (right) soccer environments. The pitch is 5m long by 4m wide. The real environment was also equipped with a motion capture (mocap) system for tracking the two robots and the ball.}
  \label{fig:sim-vs-real}
\end{figure}

Our work focuses on learning-based full-body control of humanoids for long-horizon tasks. In particular, we used deep RL to train low-cost off-the-shelf robots to play multi-robot soccer well beyond the level of agility and fluency that is intuitively expected from this type of robot. Sports like soccer showcase many of the hallmarks of human sensorimotor intelligence, which has been recognized in the robotics community, especially through the RoboCup \citep{KitanoRoboCup, RoboCupWeb} initiative. We consider a subset of the full soccer problem, and trained an agent to play simplified one-vs-one (1v1) soccer in simulation and directly deployed the learned policy on real robots (\figureref{fig:sim-vs-real}). We focused on sensorimotor full-body control from proprioceptive and motion capture observations.

The trained agent exhibited agile and dynamic movements, including walking, side stepping, kicking, fall recovery, ball interaction and more, and composed these skills smoothly and flexibly. The agent discovers surprising strategies which make more use of the full capabilities of the system than scripted alternatives, and which we may not even conceive of. An example of this is the emergent turning behavior, in which the robot pivots on the corner of a foot and spins, which would be challenging to script, and outperforms the more conservative baseline (see \nameref{sec:baseline_comparison}). One further problem in robotics generally and robot soccer in particular is the fact that optimal behaviors are often context dependent in a way that can be hard to predict and manually implement. We demonstrate that the learning approach can discover behaviors which are optimized to the specific game situation. Examples include context-dependent agile skills such as kicking a moving ball, emergent tactics such as subtle defensive running patterns, and footwork which adapts to the game situation such as taking shorter steps when approaching an attacker in possession of the ball compared to when chasing a loose ball (see \nameref{sec:behavior_analysis}). The agent learned to make predictions about the ball and the opponent, to adapt movements to the game context, and to coordinate them over long timescales for scoring, while being reactive to ensure dynamic stability. Our results also indicate that with appropriate regularization, domain randomization, and noise injected during training, safe sim-to-real transfer is possible even for low-cost robots. Example behaviors and game-play can be seen in the accompanying movies and on the website:  \url{https://sites.google.com/view/op3-soccer}.

\begin{figure}[H]
  \centering
  \includegraphics[width=\textwidth]{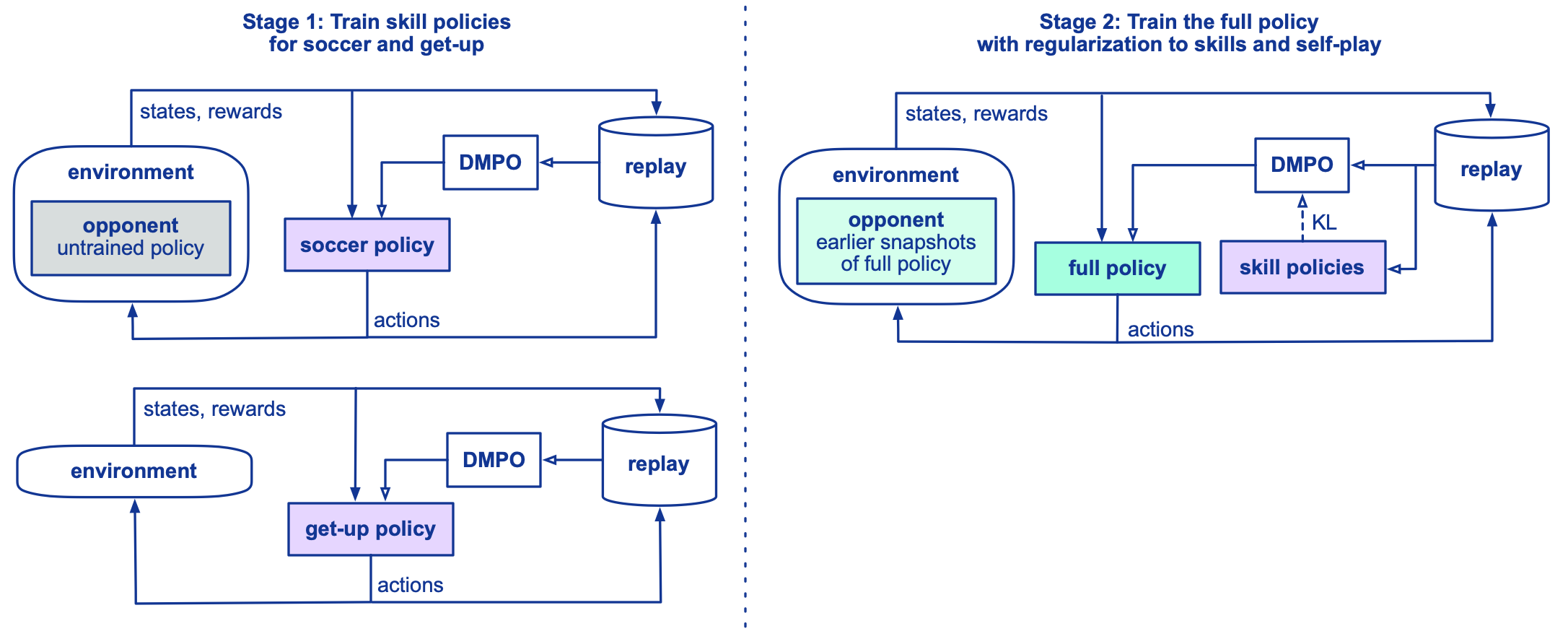}
  \caption{\textbf{Agent training setup.} We trained agents in two stages. In the first stage (left), we train a separate soccer skill and get-up skill (\nameref{sec:teacher_training}). In the second stage (right), we distill these two skills into a single agent that can both get up from the ground and play soccer (\nameref{sec:teacher_distillation_selfplay}). The second stage also incorporates self-play: the opponent is uniformly randomly sampled from saved policy snapshots from earlier in training. We found that this two-stage approach leads to qualitatively better behavior and improved sim-to-real transfer, compared to training an agent from scratch for the full 1v1 soccer task.}
  \label{fig:learning}
\end{figure}

Our training pipeline consists of two stages: In the first stage, we trained two skill policies; one for getting up from the ground, and another for scoring a goal against an untrained opponent. In the second stage, we trained agents for the full 1v1 soccer task by distilling the skills and using multi-agent training in a form of self-play, where the opponent was drawn from a pool of partially-trained copies of the agent itself. Thus in the second stage, the agent learned to combine previously learned skills, refine them to the full soccer task, and predict and anticipate the opponent's behavior. We utilized a small set of shaping rewards, domain randomization, and random pushes and perturbations to improve exploration and to facilitate safe transfer to real robots. An overview of the learning method is shown in \figureref{fig:learning}, \movref{\movoverview}, discussed in \nameref{sec:method}.

We found that pre-training separate soccer and get-up skills was the minimal set needed to succeed at the task. Learning end-to-end without separate soccer and get-up skills resulted in two degenerate solutions depending on the exact set-up: converging to either a poor locomotion local optimum of rolling on the ground or focusing on standing upright and failing to learn to score; see \nameref{sec:ablations} for more details. Using a pre-trained get-up skill simplified the reward design and exploration problem, and avoids poor locomotion local optima. More skills could be used, but we use pre-trained skills only when necessary since using a minimal set of skills allows emergent behaviors to be discovered by the agent and optimized for specific contexts (highlighted in \nameref{sec:baseline_comparison} and \nameref{sec:behavior_analysis}), rather than learning to sequence pre-specified behaviors.
% Combined environment and results section
\section*{Results}\label{sec:results}

\begin{figure}[H]
  \centering
  \includegraphics[width=\textwidth]{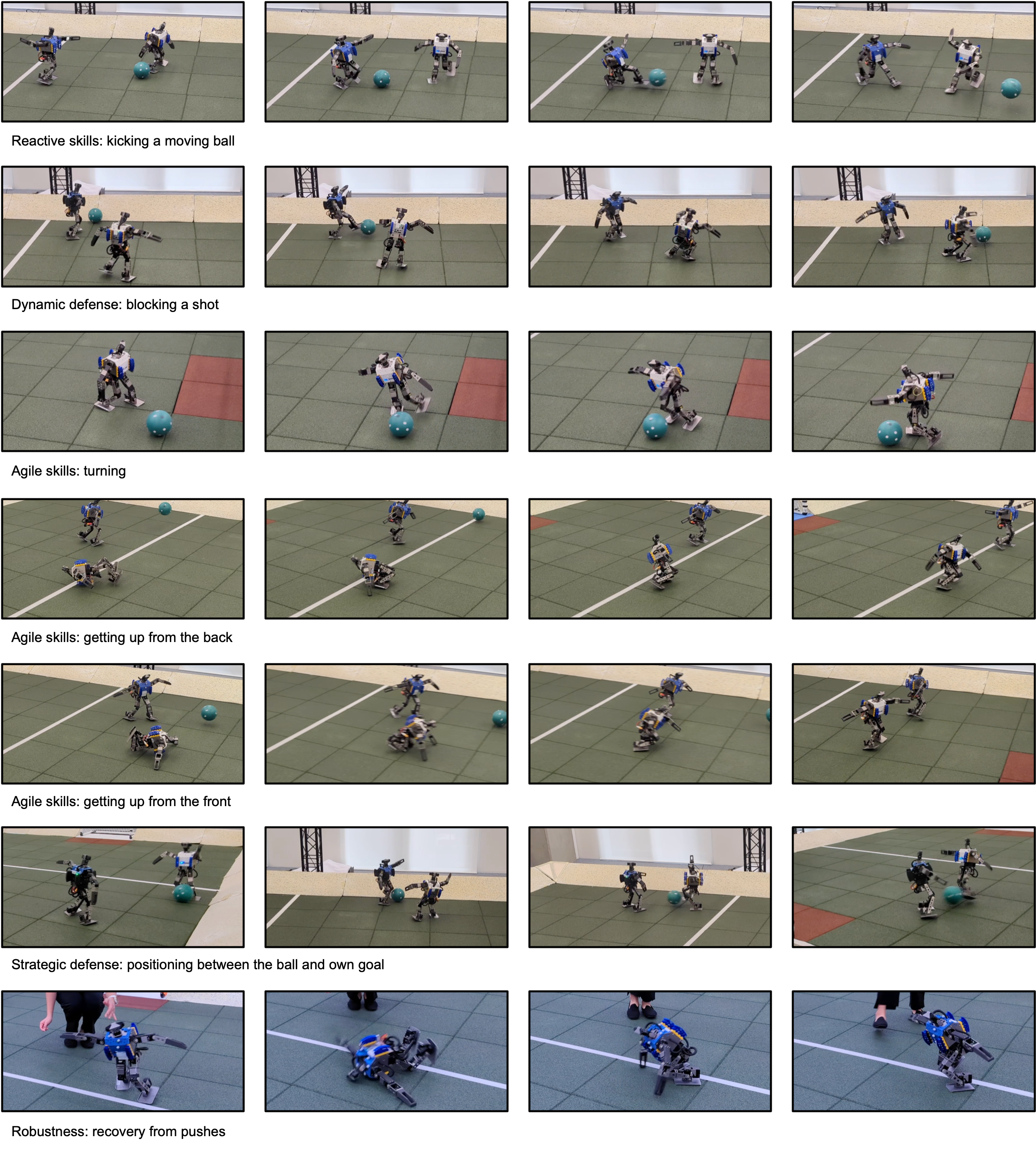}
  \caption{\textbf{Gallery of robot behaviors.} Each row gives an example of a type of behavior that is observed when trained policies are deployed on real robots.}
  \label{fig:behaviors}
\end{figure}

We evaluated the agent in a 1v1 soccer match on physical Robotis OP3 miniature humanoid robots \citep{robotisOP3}, and analyzed the emergent behaviors. We isolated certain behaviors (walking, turning, getting up, and kicking), compared them to corresponding scripted baseline controllers (\nameref{sec:baseline_comparison}), and qualitatively analyzed the behaviors in a latent space (\nameref{sec:behavior_embeddings}). To assess reliability, gauge performance gaps between simulation and reality, and study the sensitivity of the policy to game state, we also investigated selected set pieces
(\nameref{sec:behavior_analysis}). Finally, we investigated the agent's sensitivity to the observations of the ball, goal, and the opponent using value function analyses (\nameref{sec:value_function_analysis}).

Selected extracts from the 1v1 matches can be seen in \figureref{fig:behaviors} and \movref{\movbehaviors}. The agent exhibited a variety of emergent behaviors, including agile movement behaviors such as getting up from the ground, quick recovery from falls, running, and turning; object interaction such as ball control and shooting, kicking a moving ball, and blocking shots; and strategic behaviors such as defending by consistently placing itself between the attacking opponent and its own goal, and protecting the ball with its body. During play the agents transitioned between all of these behaviors fluidly.

\subsection*{Comparison to Scripted Baseline Controllers}
\label{sec:baseline_comparison}
Certain key locomotion behaviors, including getting up, kicking, walking, and turning are available for the OP3 robot \citep{op3-driver}, and we used these as baselines. The baselines are parameterized open-loop trajectories. For example, the walking controller, which can also perform turning, has tunable step length, step angle, step time, and joint offsets. We optimized the behaviors by performing a grid search over step length (for walking), step angle (for turning), and step time (for both) on a real robot. We adjusted the joint offsets where necessary to prevent the robot from losing balance or the feet colliding with each other. The kick and the get-up controllers are specifically designed for this robot and have no tunable parameters. To measure how well the learned deep RL agent performed on these key behaviors, we compared it both quantitatively and qualitatively against the baselines. See \movref{\movbaselinecomparison} for an illustration of the baselines and a side-by-side comparison with the corresponding learned behaviors. 

\begin{table}[H]
    \centering
    \small
    \begin{tabular}{lcccc}
        \toprule
        \multicolumn{5}{c}{\textbf{Set Piece Performance}} \\
        \toprule
         & \multicolumn{2}{c}{\textbf{Scoring success rate}} & \multicolumn{2}{c}{\textbf{Mean time to first touch}} \\
         & \textbf{Sim env.} & \textbf{Real env.} & \textbf{Sim env.} & \textbf{Real env.} \\
        \toprule
        Get-up and shoot & 0.70 (0.07) & 0.58 (0.07) & 4.6s (0.16s) & 4.7s (0.14s)  \\
        \bottomrule
        \toprule
        \multicolumn{5}{c}{\textbf{Behavior Analysis}} \\
        \toprule
         & \textbf{Walking speed} & \textbf{Turning speed} & \textbf{Get-up time} & \textbf{Kicking speed} \\
         & \textbf{mean (std. err.)} & \textbf{mean (std. err.)}  & \textbf{mean (std. err.)} & \textbf{mean (std. err.)} \\
        \toprule
        Scripted baseline & 0.20m/s (0.005m/s) & 0.71rad/s (0.04rad/s) & 2.52s (0.006s) & 2.07m/s (0.05m/s) \\
        Learned policy & \multirow{2}{*}{0.57m/s (0.003m/s)} & \multirow{2}{*}{2.85rad/s (0.19rad/s)} & \multirow{2}{*}{0.93s (0.12s)} & \multirow{2}{*}{2.02m/s (0.26m/s)} \\
        (real robot) \\
        With run up & - & - & - & 2.77m/s (0.11m/s) \\
        Learned policy & \multirow{2}{*}{0.51m/s (0.01m/s)} & \multirow{2}{*}{3.19rad/s (0.12rad/s)} & \multirow{2}{*}{0.73s (0.01s)} & \multirow{2}{*}{2.12m/s (0.07m/s)} \\
        (simulation) \\
        \bottomrule
    \end{tabular}
    \caption{Performance in the get-up-and-shoot set piece and at specific behaviors. Performance in simulation and on the real robot are compared, and the learned behavior is compared to the scripted baseline at the four behaviors. Values in braces are standard errors. The learned policy's mean kicking power was roughly equivalent to the scripted behavior from a standing pose, but with an additional run up approach to the ball the learned policy achieved a more powerful kick.}
    \label{table:Analysis}
\end{table}

Details of the comparison experiments are given in \aref{app:baseline_comparison_details}, and results are given in \tableref{table:Analysis}. The learned policy performed better than the specialized manually-designed controller: it walked \SI{181}{\percent} faster, turned \SI{302}{\percent} faster, and took \SI{63}{\percent} less time to get up. When initialized near the ball, the learned policy kicked the ball with \SI{3}{\percent} less speed; both achieved a ball speed of around \SI{2}{\m/\s}. However, with an additional run-up approach to the ball, the learned policy's mean kicking speed was \SI{2.8}{\m/\s} (\SI{34} {\percent} faster than the scripted controller) and the maximum kicking speed across episodes was \SI{3.4}{\m/\s}. As well as outperforming the scripted get-up behavior, in practice the learned policy also reacts to prevent falling in the first instance, see \supmovref{\movpushes}.

Close observation of the learned policy (as shown in \movref{\movbaselinecomparison}, \movref{\movslomoturning}, \supfigureref{fig:learned_gait}, and \supfigureref{fig:scripted_gait}) reveals it has learned to use a highly dynamic gait.  Unlike the scripted controller, which centers the robot's weight over the feet and keeps the foot plates almost parallel to the ground, the learned policy leans forward and actively pushes off from the edges of the foot plate at each step, landing on the heels.

The forward running speed of \SI{0.57}{\m/\s} and turning speed of \SI{2.85}{\radian/\s} achieved by the learned policy on the real OP3 compare favorably with the values of \SI{0.45}{\m/\s} and \SI{2.01}{\radian/\s} reported in \citep{BestmannBipedalWalking}, on simulated OP3 robots.  This latter work optimized parametric controllers of the type used by top-performing RoboCup teams. It was evaluated only in the RoboCup simulation, and the authors note that the available OP3 model featured unrealistically powerful motors. Although those results are not precisely comparable due to methodological differences, this gives an indication of how our learned policy compares to parameterized controllers implemented on the OP3 in simulation.

\subsection*{Behavior Embeddings}
\label{sec:behavior_embeddings}
A motivation for adopting end-to-end learning for the 1v1 soccer task was to obtain a policy that could blend many behaviors continuously, to react smoothly during play. To illustrate how the learned policy does this, we took inspiration from the analysis of Drosophila motion \citep{10.7554/eLife.46409}, and treated the motions as paths through 20-dimensional joint space. We used Uniform Manifold Approximation and Projection (UMAP) \citep{2018arXivUMAP} to approximately embed these paths into three-dimensional space to better visualize the behaviors.

\begin{figure}[H]
  \centering
  \includegraphics[width=\textwidth]{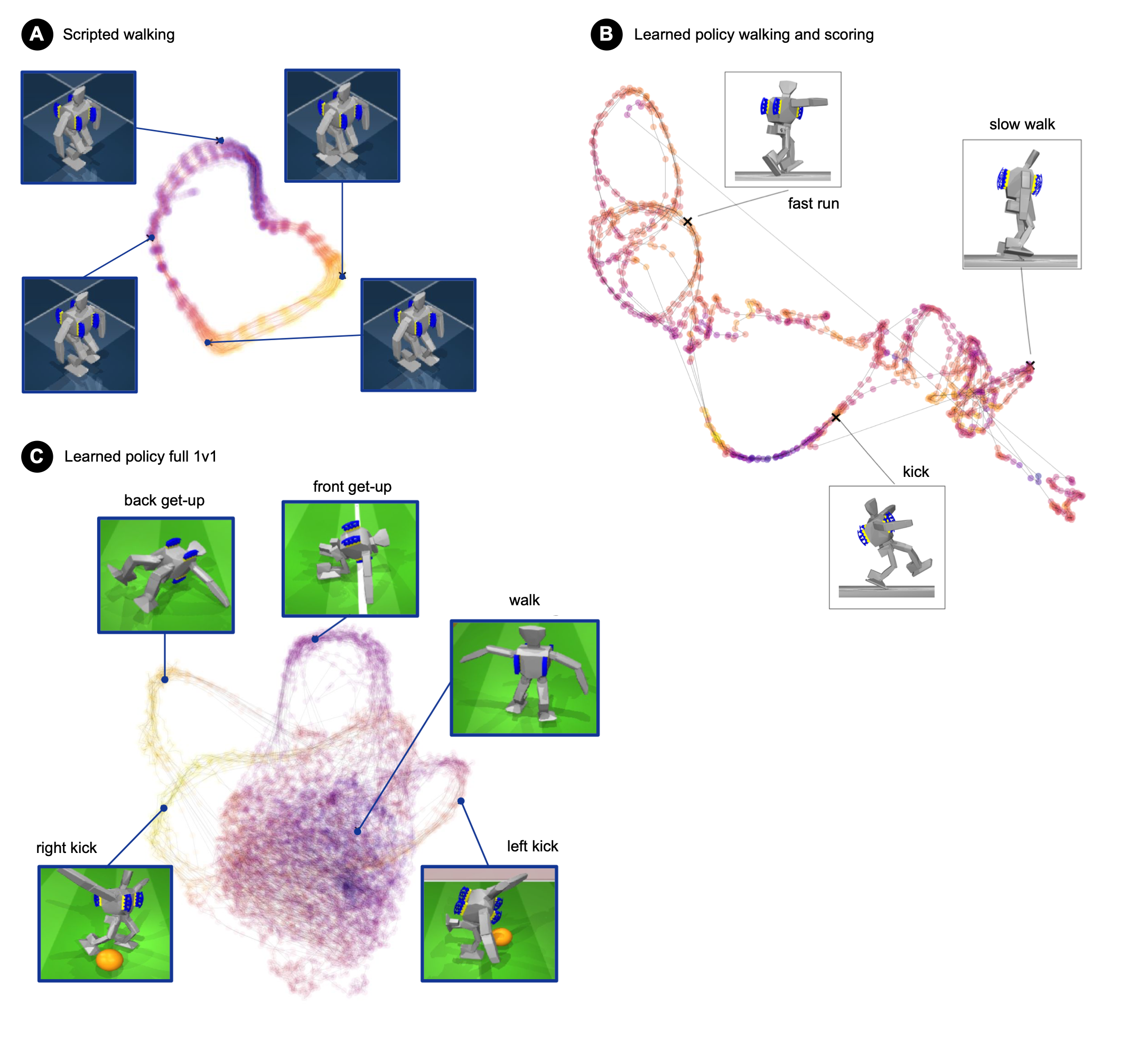}
  \caption{\textbf{Joint angle embeddings.} Embedding of the joint angles recorded while executing different policies, as described in \nameref{sec:behavior_embeddings}. \textbf{A:} The embedding for the scripted baseline walking policy. \textbf{B:} The embedding for the soccer skill. \textbf{C:} The embedding for the full 1v1 agent.}
  \label{fig:umap-full}
\end{figure}

\figureref{fig:umap-full} compares the scripted and learned embeddings. The scripted walking controller is based on sinusoidal motions of the end effectors, and this periodicity means that the gait traces a cyclic path through joint space.  This topological structure appears in the UMAP embedding, with the angular coordinate around the circle defined by the phase within the periodic gait (\figureref{fig:umap-full}A). In contrast, embeddings of short trajectories from the learned policy reveal richer variation (\figureref{fig:umap-full}B).  Footsteps still appear as loops, but the gaits are no longer exactly periodic, so the embedded trajectories are helical rather than cyclic. Different conditions also elicit different gaits, such as fast running and slower walking, and these embed as different components.  A kick common to these trajectory snippets appears as a smooth arc.

Finally, \figureref{fig:umap-full}C shows embeddings for long episodes of 1v1 play. The figure reveals a wide range of different cyclic gaits that map in a dense ball in this low-dimensional embedding space. However, kicking and getting up shows much less variation, resulting in four distinct loops. This could be a result of the regularization to the scripted get-up controller: even though the final policy could in principle converge to any behavior, regularization steers learning towards a specific way of getting up. On the other hand, kick motion is dominated by swinging one of the legs as fast as possible, allowing less room for variations.

\subsection*{Behavior Analysis}
\label{sec:behavior_analysis}

\paragraph{Reliability and Sim-to-Real Analysis:}

To gauge the reliability of the learned agent, we designed a get-up-and-shoot set piece, implemented in both the simulation (training) environment and the real environment. This set piece is a short episode of 1v1 soccer in which the agent must get up from the ground and score within 10 seconds; see \aref{app:set_piece_details} for full details.

\begin{figure}[H]
  \centering
  \includegraphics[width=0.95\textwidth]{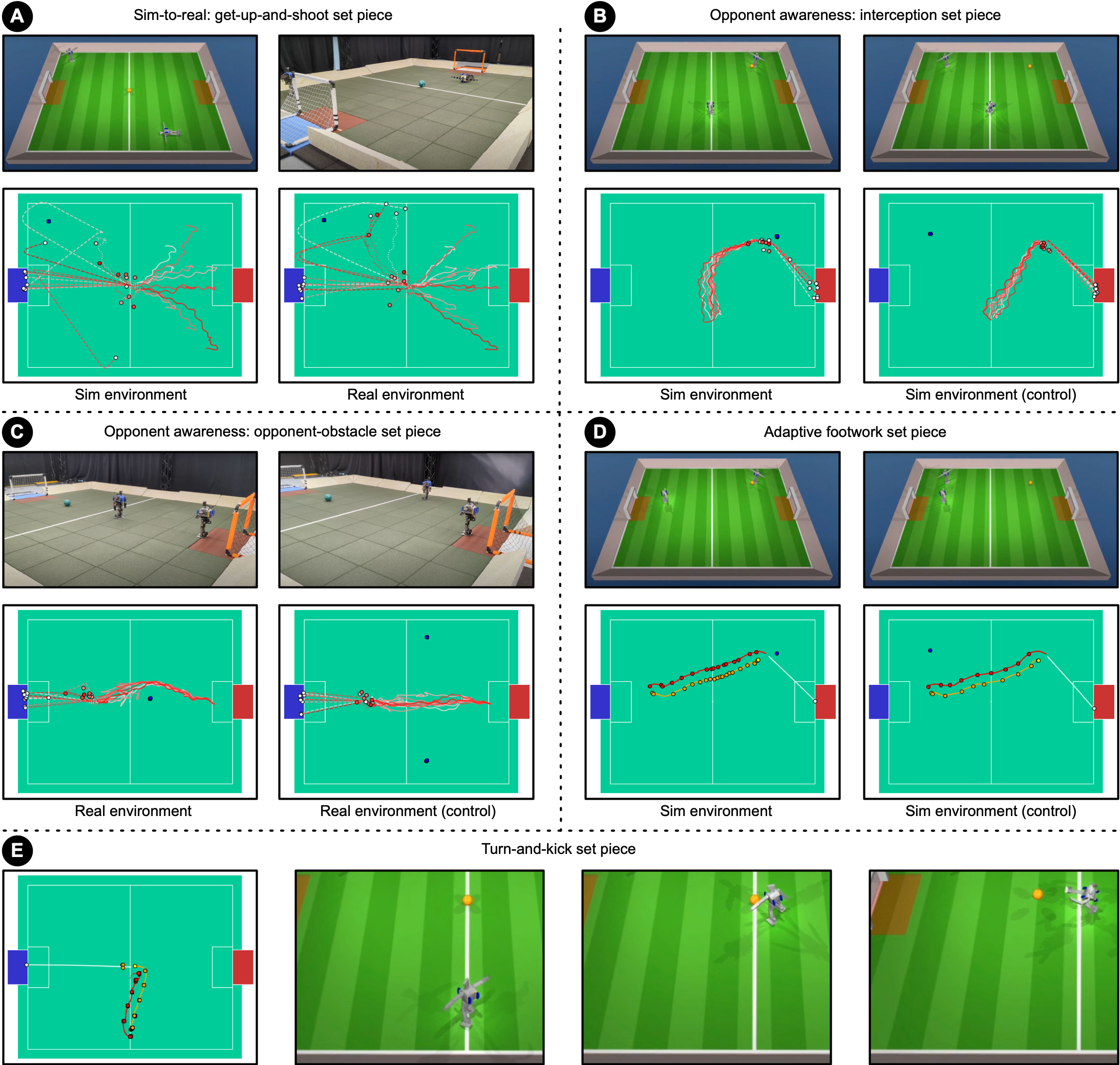}
  \caption{\textbf{Behavior analysis.} \textbf{A-C: Set pieces.} Top rows: Example initializations for the set piece tasks in sim and on the real robot. Second rows: Overlayed plots of the 10 trajectories collected from the set piece experiments showing the robot trajectory before kicking (solid lines), after kicking (dotted lines), the ball trajectory (dashed lines), final ball position (white circle), final robot position (red-pink circles) and opponent position (blue circle). Each red-pink shade corresponds to one of the 10 trajectories.
  \textbf{D: Adaptive footwork set piece.} Right foot trajectory (orange), left foot trajectory (red), ball trajectory (white), point of the kick (yellow), and footsteps highlighted with dots.
  \textbf{E: Turn-and-kick set piece.} Right 3: a sequence of frames from the set piece. Left: a plot of the footsteps from the corresponding trajectory. The agent turned, walked approximately \SI{2}{\m}, turned, kicked, and finally balanced using 10 footsteps. Please refer to \nameref{sec:behavior_analysis} for a discussion of these results.}
  \label{fig:set_piece_analysis}
\end{figure}

We played 50 episodes of the set piece each in the simulation and the real environment. In the real environment, the robot scored 29 out of 50 (\SI{58}{\percent}) of goals, and was able to get up from the ground and kick the ball every time. In simulation, the agent scored more consistently, scoring 35 out of 50 (\SI{70}{\percent}) of goals. This indicates a drop in performance due to transfer to the real environment, but the robot was still able to reliably get up, kick the ball, and score the majority of the time. Results are given in \figureref{fig:set_piece_analysis}A and \tableref{table:Analysis}.

To further gauge the sim-to-real gap, we also analyzed the four behaviors discussed in \nameref{sec:baseline_comparison} (walking, turning, getting up and kicking) in simulation and compared the results to those obtained using the real OP3. Results are shown in Table~\ref{table:Analysis}: when implemented on the real robot the learned policy walked (\SI{13}{\percent}) faster, turned (\SI{11}{\percent}) more slowly, took (\SI{28}{\percent}) more time to get up, and kicked (\SI{5}{\percent}) more slowly than when implemented in simulation. These results indicate no extreme sim-to-real gap in the execution of any behavior. The gap with the baseline behavior performance is substantially larger, for instance. The turning behavior is highly optimized (pivoting on a corner of the foot) and can be seen both in simulation and on the real robot in \movref{\movslomoturning}.

\paragraph{Opponent Awareness:}

To gauge the learned behavior's reaction to the opponent in a controlled setting, we implemented interception and opponent-obstacle set pieces in the simulation and real environments respectively.

The interception set piece is an 8 second episode of 1v1 soccer in which the opponent is initialized in possession of the ball, and remains stationary throughout the episode. In 10 trials the agent first walked to the path between the ball and its own goal, to block a potential shot from the opponent, before turning towards the ball and approaching the ball and opponent. This type of defensive behavior is manually implemented in some RoboCup teams to defend the goal against an attacker with possession \citep{BHumanCodeRelease2019}. Our learned policy, in contrast, discovered this tactic ``on its own'' by optimizing for task reward (which includes minimizing opponent scoring), rather than via manual specification. In 10 control trials the opponent was initialized away from the ball, and in these situations the agent approached the ball directly.

The opponent-obstacle set piece is a 10 second episode of 1v1 soccer in which the opponent is initialized midway between the ball and the agent, \SI{1.5}m from each, and remains stationary throughout the episode, and the ball is \SI{1.5}{\m} from the goal. In 10 trials the agent walked around the opponent every time in order to reach the ball, and scored in 9 out of 10 trials. In 14 control trials, the opponent was initialized to either side of the pitch, not obstructing the agent's path to the ball, and in this case the agent approached the ball directly, and scored in 13 of the 14 trials.

These results demonstrate that behaviors which subtly adapt to the position of the opponent emerge during training, resulting in a policy that is optimized for specific contexts. Results and initial configurations are given in \figureref{fig:set_piece_analysis}B-C.

\paragraph{Adaptive Footwork:}

In the adaptive footwork set piece, the opponent is initialized in possession of the ball and remains stationary throughout the episode, and the agent is placed in a defensive position. In 10 trials the agent took an average of 30 short footsteps to approach the attacker and ball. In comparison, in 10 control trials, in which the opponent is positioned away from the ball, the agent took an average of 20 longer strides as it rushed to the ball. The particular short-stepping tactic discovered by the agent is reminiscent of human 1v1 defensive play in which short quick steps are preferred to long strides in order to maximize reactivity. Although the reason for using short footsteps is unclear in our environment (it could be, for example, that the agent takes extra care to stay on the path between ball and goal and so moves more slowly), this result demonstrates that the agent adapts its gait to the specific context of the game. Results are illustrated in \figureref{fig:set_piece_analysis}D.

To further demonstrate the efficiency and fluidity of the discovered gait, we analyzed the footstep pattern of the agent in a turn-and-kick set piece. In this task the agent is initialized near the sideline and facing parallel with it, with the ball in the center. The natural strategy for the agent to score is therefore to turn to face the ball, then walk around the ball and turn again to kick, in a roughly mirrored ``S'' pattern. As seen in \figureref{fig:set_piece_analysis}E, the agent achieved this with only 10 footsteps. Turning, walking and kicking were seamlessly combined, and the agent adapted its footwork by taking a single penultimate shorter step to position itself to turn and kick.

\subsection*{Value Function Analysis}
\label{sec:value_function_analysis}
\begin{figure}[H]
  \centering
  \includegraphics[width=\textwidth]{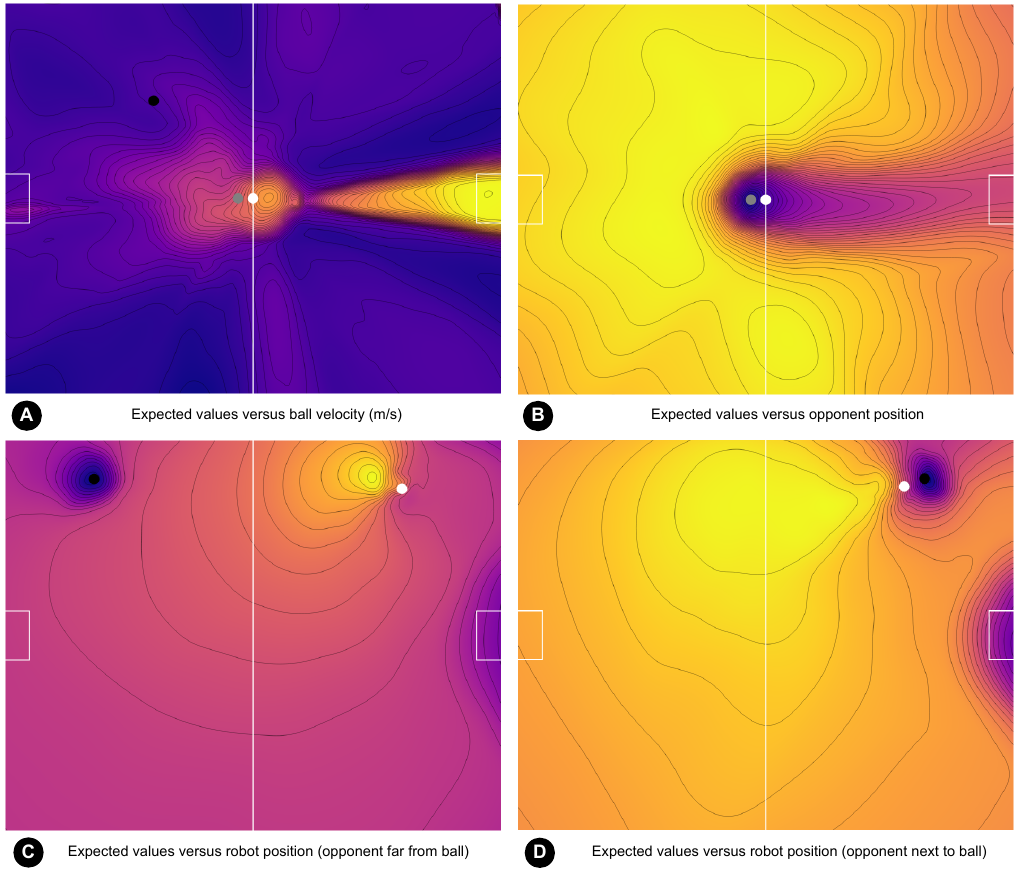}
  \caption{\textbf{Critic's predicted values.} Projections of the expected values learned by the critic, in selected game states.  (In each figure, the ball is marked in white, opponent in black, the agent in grey, and brighter colors indicate preferred states.)  \textbf{A:} Varying the ball (x,y) velocity (instantaneous velocity in m/s, mapped to the position after one second). High value states are those when the ball is either traveling toward the goal, or remaining near the agent.  \textbf{B:} Varying the opponent position around the pitch. The predicted value is lower when the opponent is located between the ball and the target goal. \textbf{C:} Varying the agent position, when the opponent is far from the ball.  (High value at positions near where the agent shoot, low value around the opponent reflecting the interference penalty).  \textbf{D:} Varying the agent position, when the opponent is close to the ball. (The value function has higher value ridges at locations blocking the opponent's shot.)}
  \label{fig:value_fn_analysis}
\end{figure}

Next, we investigated the learned value function in several setups, to understand what game states the agent perceives as advantageous versus disadvantageous.  For \figureref{fig:value_fn_analysis}A, we created a synthetic observation with the robot located centrally on the court at coordinates $(0, 0)$, facing toward the opponent goal with the ball \SI{0.1}{\m} in front of it at location $(0.1, 0)$. The opponent was placed to one side, at location $(0, 1.5)$. The plot shows the predicted value as a function of the ball's velocity vector; high value is assigned to ball velocities directed toward the goal, and also to ball velocities consistent with keeping the ball in the area under the agent's control.

A similar analysis shows that the value function also captures that the opponent impedes scoring (\figureref{fig:value_fn_analysis}B).  We used the same positions for the agent and ball, and plot the predicted value as a function of the opponent position.  States in which the opponent is positioned between the ball and the goal have a much lower value.

Figures~\ref{fig:value_fn_analysis}C-D plot the predicted value as a function of the agent's location, in scenarios like those used for the interception behavior described above in \nameref{sec:behavior_analysis}. When the opponent is far from the ball, the agent prefers to be in a position from which it can shoot,  and contour gradients tend to point directly toward this location.  In contrast, when the opponent is near the ball, the agent's preferred positions are between the ball and the defended goal, and the rigid contours favor curved paths that guide the agent to this defensive line. This behavior was seen in the interception set piece in \figureref{fig:set_piece_analysis}B.
\section*{Discussion}
\subsection*{Comparison to Robot Learning Literature}
Reinforcement learning for robots has been studied for decades (see \citep{kober2013reinforcement,deisenroth2013survey} for an overview), but has only recently gained more popularity due to the development of better hardware and algorithms \cite{peters2008reinforcement,ibarz2021train}. In particular, high-quality quadrupedal robots have become widely available, which have been used to demonstrate robust, efficient, and practical locomotion in a variety of environments \citep{lee2020learning,miki2022learning,choi2023learning,agarwal2023legged}. For example, Lee et al. \cite{lee2020learning} applied zero-shot sim-to-real deep RL to deploy learned locomotion policies in natural environments, including mud, snow, vegetation, and streaming water. Our work similarly relies on zero-shot sim-to-real transfer and model randomization, but instead focuses on a range of dynamic motions, stability, long horizon tasks, object manipulation, and multi-agent competitive play. Indeed, the vast majority of recent work in this area relies on some form of sim-to-real transfer \citep{RadosavovicLocomotion,peng2020learning,LiBipedalJumping,kumar2021rma,rudin2022advanced,bohez2022imitate,hwangbo2019learning,rudin2022advanced,smith2023learning}, which can help to reduce the safety and data efficiency concerns associated with training directly on hardware.  A common theme is that a surprisingly small number of techniques can be sufficient to reduce the sim-to-real gap\cite{ibarz2021train,muratore2022robot}, which is also supported by our results. However, there have also been successful attempts at training legged robots to walk with deep RL directly on hardware \citep{wu2023daydreamer,haarnoja2018learning,ha2021learning,smith2022walk,bloesch2022towards}. Training on hardware can lead to better performance, but the range of behaviors that can be learned has so far been limited due to safety and data efficiency concerns. Similar to our work, prior work has shown that learned gaits can achieve higher velocities compared to scripted gaits \cite{ji2022concurrent,margolis2022rapid,jin2022high}. However, the gaits have been specifically trained to attain high speeds, instead of emerging as a result of optimizing for a higher level goal.

Quadrupedal platforms constitute the majority of legged locomotion research, but an increasing number of works consider bipedal platforms. Recent works have produced behaviors including walking and running \citep{xie2019iterative,agility2022cassie}, stair climbing \citep{siekmann2021blind}, and jumping \citep{LiBipedalJumping}. Most recent works have focused on high-quality, full-sized bipeds and humanoids, with a much smaller number \citep{mordatch2015ensemble,yu2019sim,masuda2022sim,bloesch2022towards} targeting more basic platforms whose simpler and less precise actuators and sensors pose additional challenges in terms of sim-to-real transfer. Additionally, there is a growing interest in whole body control, that is, tasks in which the whole body is used in flexible ways to interact with the environment. Examples include getting-up from the ground \citep{ma2023learning} and manipulation of objects with legs \citep{cheng2023legs,nachum2019multi}. Recently, reinforcement learning has been applied to learn simple soccer skills, including goalkeeping \citep{huang2022creating}, ball manipulation \citep{bohez2022imitate,ji2023dribblebot}, and shooting \citep{ji2022hierarchical}. These works focus on a narrower set of skills than the 1v1 soccer game, and the quadrupedal platform is inherently more stable and therefore presents an easier learning challenge.

\subsection*{Comparison to RoboCup}

Robot soccer has been a longstanding grand challenge for AI and robotics, since at least the formation of the RoboCup competition \cite{RoboCupWeb,KitanoRoboCup} in 1996, and it has also inspired our 1v1 soccer task. The OP3 robot has been used for the humanoid RoboCup league, but our environment and task are substantially simpler than the full RoboCup problem. The main differences are that we focused on 1v1 soccer instead of multi-player teams; our environment does not align with the field or ball specifications nor follow the rules of RoboCup (for example, the kick-off, player substitutions, explicit communication channels, fouls, and game length); and we use full state information rather than rely solely on vision.

The majority of successful reinforcement learning approaches to RoboCup focus on learning specific components of the system and often feature manually designed components or high-level strategies. In Simulation 2D League, reinforcement learning has been used to learn various ball handling skills  \citep{Riedmiller-karlsruhebrainstormers, TuylsMM02} and multi-agent behaviors such as defense \citep{riedmiller2009reinforcement} and ball control \citep{AB05,LNAI09-kalyanakrishnan-1,LNAI2006-shivaram}. 
One successful learning-based approach applied to the Simulation 3D League is Layered Learning \cite{stone2000layered,macalpine2018overlapping}. There, RL was used to train multiple skills, including ball control and pass selection, which were combined via a pre-defined hierarchy. Our system pre-defines fewer skills (leaving the agent to discover useful skills like kicking) and we focused on learning to combine the skills seamlessly. Additionally, RL has been used for fast running \citep{abreu2019learning, melo2021learning}, but compared to our work, the learned behaviors were not demonstrated on hardware. 
A smaller set of works has focused on applying RL to real robots. Policy gradient methods have been used to optimize parameterized walking \citep{LNAI2006-manish} and kicking \citep{LNAI10-hausknecht} on a quadrupedal robot for the RoboCup Four-Legged League. Riedmiller et al.  \citep{riedmiller2009reinforcement} applied RL to learn low-level motor speed control, as well as a separate dribbling controller for the wheeled Middle Size League. Simulation grounding by sim-to-real transfer for humanoids has also been investigated by Farchy et al. \citep{AAMAS13-Farchy}. However, they focus on learning the parameters of a manually designed walk engine, whereas we learn a neural network policy to output joint angles for the full soccer task directly.

\subsection*{Limitations}
Our work provides a step towards practical use of deep RL for agile control of humanoid robots in a dynamic multi-agent setting. However, there are several topics that could be addressed further. First, our learning pipeline relies on some domain-specific knowledge and domain randomization, as is common in the robot learning literature \citep{lee2020learning, ibarz2021train, kober2013reinforcement, deisenroth2013survey, muratore2022robot}. Domain-specific knowledge is used for reward function design and for training the get-up skill, which requires access to hand-designed key poses, which can be difficult or impractical to choose for more dynamic platforms. In addition, the distillation step assumes we can manually choose the correct skill (either get-up or soccer) for each state, although a method in which the distillation target is automatically selected has been demonstrated in prior work \citep{LiuHumanoidSoccer}, which we anticipate would work in this application. Second, we do not leverage real data for transfer; instead, our approach relies solely on sim-to-real transfer. Fine-tuning on real robots or mixing in real data during training in simulation could help improve transfer and enable an even wider spectrum of stable behaviors. Third, we applied our method to a small robot and did not consider additional challenges that would be associated with a larger form factor.

Our current system could be improved in a number of ways. We found that tracking a ball with motion capture was particularly challenging: detection of the reflective tape markers is sensitive to the angle at which they face the motion capture cameras; only the markers on the upper hemisphere of the ball can be registered; and the walls of the soccer pitch can occlude the markers, especially near the corners. We believe moving away from motion capture is an important avenue for future work and discuss potential avenues for this in \nameref{sec:future_work}. We also found that the performance of the robots degraded quickly over time, mainly due to the hip joints becoming loose or the joint position encoders becoming miscalibrated; thus we needed to regularly perform robot maintenance routines. Further, our control stack was not optimized for speed. Our nominal control time step was \SI{25}{\ms}, but in practice the agent often failed to produce an action within that time. The time step was selected as a compromise between speed and consistency, but we believe that a higher control rate would result in improved performance. Finally, we did not model the servo motors in simulation, but instead approximated them with ideal actuators that can produce the exact torque requested by a position feedback controller. As a consequence, for example, we found that the agent's behaviors are very sensitive to the battery charge level, limiting the operation time per charge to 5 to 10 minutes in practice.

On the training side, we found our self-play setup sometimes resulted in unstable learning. A population-based training scheme \cite{LiuHumanoidSoccer} could have improved stability and led to better multi-agent performance. Second, our method includes several auxiliary reward terms, some of which are needed for improved transfer (for example, upright reward and knee torque penalty), and some for better exploration (for example, forward speed). We chose to use a weighted average of the different terms as the training reward, and tuned the weights via an extensive hyperparameter search. However, multi-objective RL \citep{roijers2013survey,abdolmaleki2020distributional} or constrained RL \citep{ray2019benchmarking} might be able to obtain better solutions.

\subsection*{Future Work}
\label{sec:future_work}
\textbf{Multi-agent Soccer:} An exciting direction of future work would be to train teams of two or more agents. It is straightforward to apply our proposed method to train agents in this setting. In our preliminary experiments for 2v2 soccer, we saw that the agent learned division of labor, a simple form of collaboration: if its teammate was closer to the ball, the agent did not approach the ball. However, it also learned less agile behaviors. Insights from prior work in simulation \citep{LiuHumanoidSoccer} could be applied to improve performance in this setting.

\textbf{Playing Soccer from Raw Vision:}
Another important direction for future work is learning from on-board sensors only, without external state information from a motion capture system. In comparison to state-based agents that have direct access to the ball, goal, and opponent locations, vision-based agents need to infer information from a limited history of high-dimensional egocentric camera observations, and integrate the partial state information over time, which makes the problem significantly harder \citep{TassaControlSuite}. 

As a first step, we investigated how to train vision-based agents that only use onboard RGB camera and proprioception. We created a visual rendering of our lab using a Neural Radiance Field (NeRF) model \citep{mildenhall2021nerf} based on the approach introduced by Byravan et al.~\cite{NeRF2Real}. The robot learned behaviors including ball tracking and situational awareness of the opponent and goal. See \aref{app:vision} for our preliminary results with this approach.
\section*{Materials and Methods}
\label{sec:method}

\subsection*{Environment}
\label{sec:sim-env}
We trained the agent in simulation in a custom soccer environment and then transferred to a corresponding real environment as shown in \figureref{fig:sim-vs-real}. The simulation environment uses the MuJoCo physics engine \citep{Todorov_2012} and is based on the DeepMind Control Suite \citep{tassa2020dm_control}. The environment consists of a soccer pitch that is \SI{5}{\meter} long by \SI{4}{\meter} wide, and two goals that each have an opening width of \SI{0.8}{\meter}. In both the simulated and real environments, the pitch is bordered by ramps, which ensures that the ball returns to the bounds of the pitch. The real pitch is covered with rubber floor tiles to reduce the risk of falls damaging the robots and to increase the ground friction.

The agent acts at \SI{40}{\hertz}. The action is 20-dimensional and corresponds to the joint position set points of the robot. The actions are clipped to a manually selected range (see \aref{app:environment_details}) and passed through an exponential action filter to remove high frequency components: $\ut = 0.8 \utm + 0.2 \at$, where $\ut$ is the filtered control applied to the robot at time step $t$, and $\at$ is the action output by the policy. The filtered actions are fed to PID controllers that then drive the joints (torques in simulation and voltages on the real robot) to attain the desired positions.

The agent's observations consist of proprioception and game state information. The proprioception consists of joint positions, linear acceleration, angular velocity, gravity direction, and the state of the exponential action filter. The game state information, obtained via a motion capture setup in the real environment, consists of the agent's velocity, ball location and velocity, opponent location and velocity, and location of the two goals, which enables the agent to infer its global position. The locations are given as two-dimensional vectors corresponding to horizontal coordinates in the egocentric frame, and the velocities are obtained via finite differentiation from the positions. All proprioceptive observations, as well as the observation of the agent's velocity, are stacked over the five most recent timesteps to account for delays and potentially noisy or missing observations. We found stacking of proprioceptive observations to be sufficient for learning high-performing policies, so we chose not to stack the game state (opponent, ball, and goal observations) to reduce the size of the observation space. However, including a full observation history could be important for improving the agent's ability to react and adapt to the opponent in more subtle ways. A more detailed description of the observations is given in \aref{app:environment_details}.

\subsection*{Robot Hardware and Motion Capture} \label{sec:RobotHardware}

We used the Robotis OP3 robot \citep{robotisOP3}, which is a low-cost battery-powered, miniature humanoid platform.  It is \SI{51}{\cm} tall, weighs \SI{3.5}{\kg}, and is actuated by 20 Robotis Dynamixel XM430-350-R servomotors.  We controlled the servos by sending target angles using position control mode with only proportional gain (in other words, without any integral or derivative terms).  Each actuator has a magnetic rotary encoder that provides the joint position observations to the agent. The robot also has an inertial measurement unit (IMU), which provides angular velocity and linear acceleration measurements. We found that the default robot control software was sometimes unreliable and caused nondeterministic control latency, so we wrote a custom driver that allows the agent to communicate directly and reliably with the servos and IMU via the Dynamixel SDK Python API. The control software runs on an embedded Intel Core i3 dual-core NUC with Linux.  The robot lacks GPUs or other dedicated accelerators, so all neural network computations were run on the CPU. The robot's ``head'' is a Logitech C920 web camera, which can optionally provide an RGB video stream at 30 frames per second.

The robot and ball positions and orientations were provided by a motion capture system based on Motive 2 software \citep{optitrack}.  This system uses 14 Optitrack PrimeX 22 Prime cameras mounted on a truss around the soccer pitch. We tracked the robots using reflective passive markers attached to a 3D printed ``vest'' covering the robot torso, and tracked the ball using attached reflective stickers (\figureref{fig:sim-vs-real}). The positions of these three objects were streamed over the wireless network using the VRPN protocol and made available to the robots via ROS.

We made small modifications to the robot to reduce damage from the evaluation of a wide range of prototype agent policies. We added 3D-printed safety bumpers at the front and rear of the torso, to reduce the impact of falls.  We also replaced the original sheet metal forearms with 3D-printed alternatives, with the shape based on the convex hull of the original arms, because the original hook-shaped limbs would sometimes snag on the robot's own cabling. We also made small mechanical modifications to the hip joints to spread the off-axis loads more evenly, to minimize fatigue breakages.

\subsection*{Policy Optimization}
We modeled the soccer environment as a Partially Observable Markov Decision Process (POMDP) defined by $(\cS,\cA,\mathcal{P},r,\mu_0,\gamma)$, with states $s \in \cS$, actions $a \in \cA$, transition probabilities $\mathcal{P}(\boldsymbol{s}'|\boldsymbol{s}, \boldsymbol{a})$, reward function $r(\boldsymbol{s})$, distribution over initial states $\mu_0$, and discount factor $\gamma \in [0, 1)$. At each timestep $t$, the agent observes features $\ot \triangleq \phi(\st)$, extracted from the state $\st\in\cS$ as described in \nameref{sec:sim-env}.  Actions are 20-dimensional and continuous, corresponding to the desired positions of the robot's joints. The reward is a weighted sum of $K$ reward components, $r(\boldsymbol{s}) = \sum_{k=1}^K\alpha_k\hat{r}_k(\boldsymbol{s})$; \aref{app:rewards} describes the components we use for each stage of training.

A trajectory is defined as $\traj = \left\{(\st, \at, r_{t+1})\right\}_{t=0}^{\infty}$. A policy $\pi(\boldsymbol{a}|\boldsymbol{s})$, along with the system dynamics $\mathcal{P}$ and initial state distribution $\mu_0$, gives rise to a distribution over trajectories: $\mu_\pi(\xi) = \mu_0(\sz) \prod_{t=0}^\infty \pi(\at | \st) \mathcal{P}(\stp | \st, \at)$.
The aim is to obtain a policy $\pi$ that maximizes the expected discounted cumulative reward, or return:
\begin{align}
    \label{eqn:Methods:MDPObjective}
    \cJ(\pi)\triangleq\mathbb{E}_{\traj \sim \mu_\pi}\left[ \sum^T_{t=0} \gamma^t r(\st) \right].
\end{align}
We parameterized the policy as a deep feed-forward neural network with parameters $\theta$, that outputs the mean and diagonal covariance of a multivariate Gaussian. We trained this policy to optimize \eqref{eqn:Methods:MDPObjective} using Maximum a posteriori Policy Optimization~\citep{Abdolmaleki_2018} (MPO), which is an off-policy actor-critic RL algorithm. MPO alternates between policy evaluation and policy improvement. In the policy evaluation step, the critic (or $Q$-function) is trained to estimate $Q^{\pi_\theta}(\boldsymbol{s}, \boldsymbol{a})$, which describes the expected return from taking action $\boldsymbol{a}$ in state $\boldsymbol{s}$ and then following policy $\pi_\theta$: $Q^{\pi_\theta}(\boldsymbol{s}, \boldsymbol{a})\triangleq \mathbb{E}_{\traj \sim \mu_\theta(\traj|\boldsymbol{s}_0 = \boldsymbol{s}, \boldsymbol{a}_0 = \boldsymbol{a})}[ \sum^T_{t=0} \gamma^t r(\st) ]$. We use a distributional critic~\citep{Bellamare_2017}, and refer to the overall algorithm as Distributional MPO, or DMPO. In the policy improvement step, the actor (or policy) is trained to improve in performance with respect to the $Q$-values predicted by the critic. Details of the DMPO algorithm, learning hyperparameters, and the agent architecture are given in \aref{sec:Appendix:Training:MPO}.

Note that the choice of opponent affects the transition probabilities $\mathcal{P}$, and thus the trajectory distribution $\mu_\pi$. When training the soccer skill, the opponent is fixed, but when training the full 1v1 agent in the second stage, we sampled the opponent from a pool of previous snapshots of the agent, rendering the objective both non-stationary and partially observed. In practice though, the agent was able to learn and eventually converge to a well-performing policy. Similar approaches have been explored in prior work on multi-agent deep RL~\citep{bansal2017emergent,HeinrichFictitious,lanctot2017unified}.

\subsection*{Training}
\label{sec:training}
Our training pipeline has two stages. This is because directly training agents on the full 1v1 task leads to sub-optimal behavior, as described in \nameref{sec:ablations}. In the first stage, separate skill policies for scoring goals and getting up from the ground are trained. In the second stage, the skills are distilled into a single 1v1 agent, and the agent is trained via self-play. Distillation to skills stops after the agent's performance surpasses a pre-set threshold, which enabled the final behaviors to be more diverse, fluent, and robust than simply composing the skills. Self-play provides an automatic curriculum, and expands the set of environment states encountered by the agent. \aref{app:training_curves} contains learning curves and training times for each of the skills and the full 1v1 agent.

\subsubsection*{Stage 1: Skill Training}
\label{sec:teacher_training}

\paragraph{Soccer Skill Training:}
The soccer skill is trained to score as many goals as possible. Episodes terminate when the agent falls over, goes out of bounds, enters the goal penalty area (marked with red in \figureref{fig:sim-vs-real}), the opponent scores, or a timelimit of 50 seconds is reached. At the start of each episode, the players and the ball are initialized randomly on the pitch. Both players are initialized in a default standing pose. The opponent is initialized with an untrained policy, which falls almost immediately and remains on the ground. The reward is a weighted sum over reward components. This included components to encourage forward velocity and ball interaction, to make exploration easier, as well as components to improve sim-to-real transfer and reduce robot breakages, as discussed in \nameref{sec:robot_safety}.

\paragraph{Get-Up Skill Training:} 
The get-up skill is trained using a sequence of target poses, to bias the policy towards a stable and collision-free trajectory. We used the pre-programmed get-up trajectory \citep{op3-driver} to extract three key poses for getting up from either the front or the back (see \aref{app:getupskill} for an illustration of the key poses).

We trained the get-up skill to reach any target pose interpolated between the key poses. We conditioned both the actor and critic on the target pose, which consists of target joint angles $\mathbf{p}_\mathrm{target}$ and target torso orientation $\mathbf{g}_\mathrm{target}$. The target torso orientation is expressed as the gravity direction in the egocentric frame. This is independent of the robot yaw angle (heading), which is irrelevant for the get-up task. Conditioning on the joint angles steers the agent towards collision free and stable poses, whereas conditioning on the gravity direction ensures the robot intends to stand up rather than just matching the target joint angles while lying on the ground. The robot is initialized on the ground, and a new target pose is sampled uniformly at random every 1.5 seconds on average. The sampling intervals are exponentially distributed, to make the probability of a target pose switch independent of time, in order to preserve the Markov property. The agent is trained to maximize $\hat r_{\mathrm{pose}}(\st) = -\tilde p_t\tilde g_t$, where $\tilde p_t = \nicefrac{\big(\pi - \|\mathbf{p}_\mathrm{target} - \mathbf{p}_t\|_2\big)} {\pi}$ is the scaled error in joint positions, and $\tilde g_t =\nicefrac{\big(\pi - \arccos\left(\mathbf{g}^\intercal_t\mathbf{g}_\mathrm{target}\right)\big)}{\pi}$ 
is the scaled angle between the desired and actual gravity direction. $\mathbf{p}_t$ and $\mathbf{g}_t$ are the actual joint positions and gravity direction at timestep $t$, respectively.

Conditioning the converged policy on the last key pose, corresponding to standing, makes the agent get up. We used this conditioned version as a get-up skill in the next stage of training.

\subsubsection*{Stage 2: Distillation and Self-Play}
\label{sec:teacher_distillation_selfplay}
In the second stage, the agent competes against increasingly stronger opponents, while initially regularizing its behavior to the skill policies. This resulted in a single 1v1 agent that is capable of a range of soccer skills: walking, kicking, getting up from the ground, scoring, and defending. The setup is the same as for training the soccer skill, except episodes terminate only when either the agent or the opponent scores, or after 50 seconds. When the agent is on the ground, out of bounds, or in the goal penalty area, it receives a fixed penalty per timestep and all positive reward components are ignored. For instance, if the agent is on the ground when a goal is scored, then it receives a zero for the scoring reward component. At the beginning of an episode, the agent is initialized either laying on the ground on its front, on its back, or in a default standing pose, with equal probability.

\paragraph{Distillation:}
We use policy distillation \citep{rusu2015policy,parisotto2015actor} to enable the agent to learn from the skill policies, by adding a regularization term that encourages the output of the agent's policy to be similar to that of the skills'. This approach is related to prior work that regularizes a student policy to either a common shared policy across tasks \citep{teh2017distral} or a default policy that receives limited state information \citep{Galashov2019infoasym}, as well as work on kickstarting \citep{Schmitt2018KickstartingDR} and reusing learned skills for humanoid soccer in simulation \citep{LiuHumanoidSoccer}.

Unlike most prior work, in our setting the skill policies are useful in mutually exclusive sets of states: the soccer skill is useful only when the agent is standing up, otherwise the get-up skill is more useful. Thus in each state, we regularize the agent's policy $\pi_\theta$ to only one of the two skills. We achieve this by replacing the critic's predicted Q-values used in the policy improvement step with a weighted sum of the predicted Q-values and KL-regularization to the relevant skill policy:
\begin{equation}
  \begin{cases}
    (1 - \clambdas) \; \mathbb{E}_{\boldsymbol{a} \sim \pi(\cdot|\boldsymbol{s})} \big[Q^{\pi_\theta}(\boldsymbol{s},\boldsymbol{a})\big] - \clambdas \mathrm{KL}\big(\pi_\theta(\cdot|\mathbf{s}) \| \cpis(\cdot|\mathbf{s})\big) \; & \text{ if } \boldsymbol{s} \in \mathcal{U} \\[8pt]
    (1 - \clambdag) \; \mathbb{E}_{\boldsymbol{a} \sim \pi(\cdot|\boldsymbol{s})} \big[Q^{\pi_\theta}(\boldsymbol{s},\boldsymbol{a})\big] - \clambdag \mathrm{KL}\big(\pi_\theta(\cdot|\mathbf{s}) \| \cpig(\cdot|\mathbf{s})\big) \; & \text{ if } \boldsymbol{s} \notin \mathcal{U} \; ,
  \end{cases}
  \label{eq:DistillationObjective}
\end{equation}
where $\cU$ is the set of all states in which the agent is upright.

To enable the agent to outperform the skill policies, the weights $\clambdas$ and $\clambdag$ are adaptively adjusted such that there is no regularization once the predicted Q-values are above the pre-set thresholds $\cQs$ and $\cQg$, respectively. This approach was proposed by Abdolmaleki et al.~\cite{abdolmaleki2021dime} for a similar setting and is closely related to the Lagrangian multiplier method used in constrained RL~\citep{Stooke2020}. Specifically, $\clambdas$ (or $\clambdag$) is updated by stochastic gradient descent to minimize
\begin{equation}
    c(\clambdas) = \clambdas \big(\mathbb{E}_{\xi} \big[ Q^{\pi_\theta}(\mathbf{s},\mathbf{a}) \big] - \cQs\big),
    \label{eq:KLweightobjective}
\end{equation}
using a softplus transform and clipping to enforce that $0 \leq \clambdas \leq 1$. When the agent's predicted return is less than $\cQs$, then $\clambdas$ increases to $1$, at which point the agent effectively performs behavioral cloning to the soccer skill. Once the agent's predicted return surpasses $\cQs$, then $\clambdas$ decreases to $0$, at which point the agent learns using pure RL on the soccer training objective. This enabled the agent to improve beyond any simple scheduling of the skill policies; our agents learned effective transitions between the two skills and finetuned the skills themselves.

\paragraph{Self-Play:}
The performance and learned strategy of an agent depends on its opponents during training. The soccer skill plays against an untrained opponent, and thus this policy had limited awareness of the opponent. To improve agents' high level game play, we used self-play, where the opponent is drawn from a pool of partially trained copies of the agent itself~\cite{bansal2017emergent,HeinrichFictitious,lanctot2017unified}. Snapshots of the agent are regularly saved, and the first quarter of the snapshots is included in the pool, along with an untrained agent. We found that using the first quarter, rather than all snapshots, improved stability of training by ensuring that the performance of the opponent improves slowly over time. In our experiments, self-play training led to agents that were agile and defended against the opponent scoring.

Playing against a mixture of opponents results in significant partial observability due to aliasing with respect to the opponent in each episode. This can cause significant problems for critic learning, since value functions fundamentally depend upon the opponent's strategy and ability~\cite{LiuSoccer}. To address this, we condition the critic on an integer identification of the opponent.

\subsection*{Sim-to-Real Transfer}
\label{sec:transfer}
Our approach relies on zero-shot transfer of trained policies to real robots. This section details the approaches we took to maximize the success of zero-shot transfer: we reduced the sim-to-real gap via simple system identification, improved the robustness of the policies via domain randomization and perturbations during training, and included shaping reward terms to obtain behaviors that are less likely to damage the robot.

\subsubsection*{System Identification}
We identified the actuator parameters by applying a sinusoidal control signal of varying frequencies to a motor with a known load attached to it, and optimized over the actuator model parameters in simulation to match the resulting joint angle trajectory. For simplicity, we chose a position controlled actuator model with torque feedback and with only damping (\SI{1.084}{\N\m\per(\radian\per\s)}), armature (\SI{0.045}{\kg\,\m\squared}), friction (0.03), maximum torque (\SI{4.1}{\N\m}), and proportional gain (\SI{21.1}{\N/\radian}) as free parameters. The values in parentheses correspond to the final values after applying this process. This model does not exactly correspond to the servos' operating mode, which controls the coil voltage instead of output torque, but we found it matched the training data sufficiently well. We believe this is because using a position control mode hides model mismatch from the agent by applying fast stabilizing feedback at a high frequency. Indeed, we also experimented with direct current control, but found the sim-to-real gap was too large, which caused zero-shot transfer to fail. We expect that the sim-to-real gap could be further reduced by considering a more accurate model. 

\subsubsection*{Domain Randomization and Perturbations}
\label{sec:domain_randomization}
To further improve transfer, we applied domain randomization and random perturbations during training. Domain randomization helps overcome the remaining sim-to-real gap and the inherent variation in dynamics across robots, due to wear and other factors such as battery state. We selected a small number of axes to vary, since excess randomization could result in a conservative policy, which would reduce the overall performance. Specifically, we randomized the floor friction (0.5 to 1.0) and joint angular offsets ($\pm$ 2.9\degree); varied the orientation (up to 2\degree) and position (up to \SI{5}{\mm}) of the IMU; and attached a random external mass (up to \SI{0.5}{\kg}) to a randomly chosen location on the robot torso. We also added random time delays (10 to \SI{50}{\ms}) to the observations to emulate latency in the control loop. These domain randomization settings were re-sampled at the beginning of each episode and then kept constant for the whole episode. In addition to domain randomization, we found that applying random perturbations to the robot during training substantially improved the robustness of the agent, leading to better transfer. Specifically, we applied an external impulse force of 5 to \SI{15}{\N\m} lasting for 0.05 to \SI{0.15}{\s}, to a randomly selected point on the torso  every 1 to \SI{3}{\s}.

Zero-shot transfer did not work for agents trained without domain randomization and perturbations: when deployed on physical robots, these agents would fall over with every one or two steps, and were unable to score.

\subsubsection*{Regularization for Safe Behaviors}
\label{sec:robot_safety}
We limited the range of possible actions for each joint to allow sufficient range of motion while minimizing the risk of self-collisions (\tableref{table:joint_limits} in \aref{app:environment_details}). We also included two shaping reward terms to improve sim-to-real transfer and reduce robot breakages (see \tableref{table:ShapingRewards} in \aref{app:training_details}). In particular, we found that highly dynamic gaits and kicks often led to excessive stress on the knee joints from the impacts between the feet and the ground or the ball, which caused gear breakage. We mitigated this by regularizing the policies via a penalty term to minimize the time integral of torque peaks (thresholded above \SI{5}{\N\m}) as calculated by MuJoCo for the constraint forces on the targeted joints. In addition to the knee breakages, we noticed the agent would often lean forward when walking. This made the gait faster and more dynamic, but when transferred to a real robot, it would often cause the robot to loose balance and fall forward. To mitigate this effect we added a reward term for keeping an upright pose within the threshold of  11.5\degree. Incorporating these two reward components led to robust policies for transfer that rarely broke knee gears, and performed well at scoring goals and defending against the opponent.

\subsection*{Ablations}
\label{sec:ablations}
We ran ablations to investigate the importance of regularization to skill policies and using self-play, described below. We also ran ablations on the reward components (\aref{app:reward_ablations}).

\subsubsection*{Importance of Regularization to Skill Policies}
\begin{figure}[H]
  \centering
  \includegraphics[width=\textwidth]{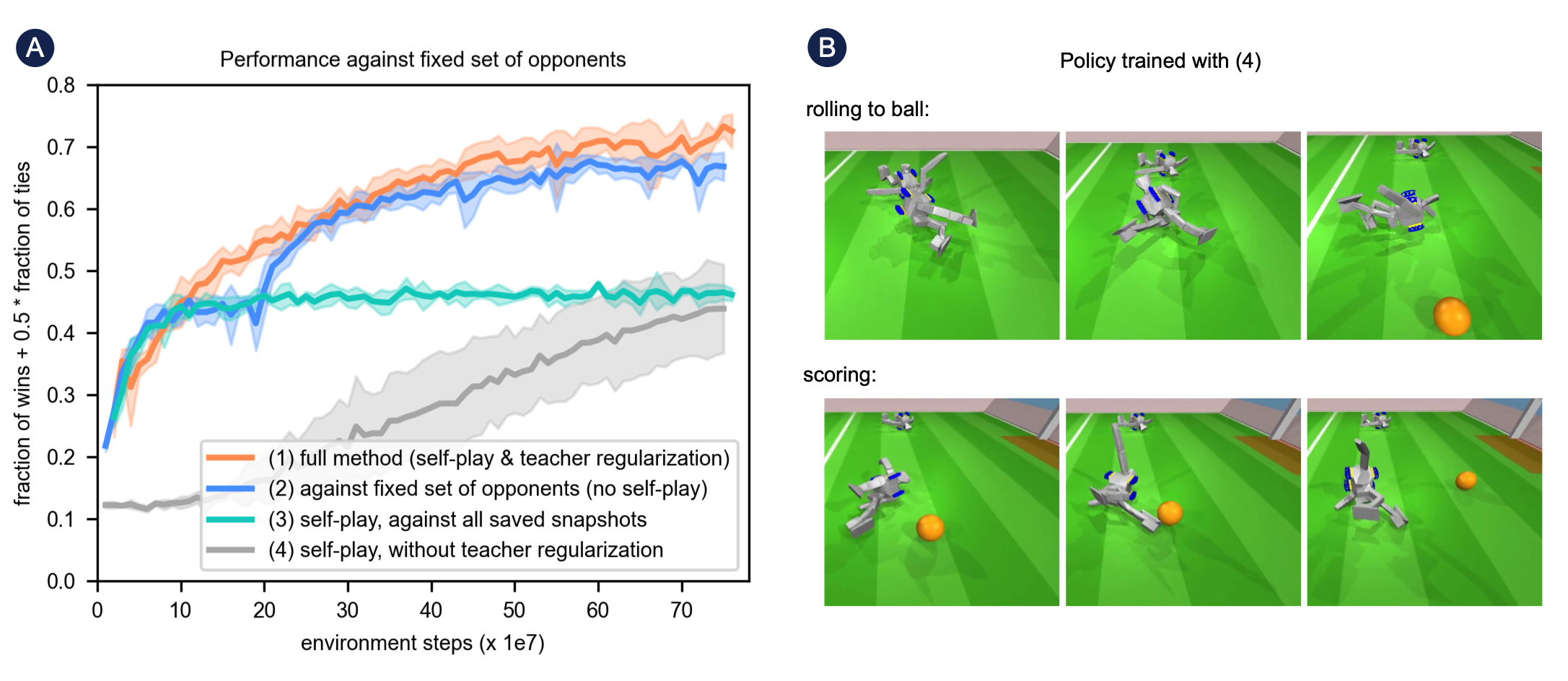}
  \caption{\textbf{Comparison against ablations.} \textbf{A:} The plot compares our full method (1) against ablations: at regular intervals across training, each policy was evaluated against a fixed set of six opponents, playing 100 matches against each. The first player to score in the episode wins the match; if neither player scores within 50 seconds, then it is a tie. The y-axis corresponds to the fraction of wins (with a draw counting as 1/2 of a win), averaged across the six opponents. We ran five seeds per method, and the error bars show 95\% confidence intervals. Our full method (1) includes both self-play and regularization to pre-trained skills. In (2), instead of self-play, agents are trained against the same fixed set of six opponents as in the evaluation. Interestingly, agents trained with self-play in (1) performed better when evaluated against this fixed set of opponents, despite not playing directly against them during training. In (3), self-play consists of training against all saved snapshots, rather than the first quarter. This led to less stable learning and converged to worse performance. In (4), agents are trained without regularization to skills and without reward shaping. \textbf{B:} Two sequences of frames taken from an agent trained with (4). These agents did not learn to get up from the ground; instead, they learned to score by rolling to the ball and knocking it into the goal.}
  \label{fig:ablations}
\end{figure}

First, we trained agents without regularization to skill policies. When we gave agents a sparse reward, only for scoring or conceding goals, they learned a local optimum of rolling to the ball and knocking it into the goal with a leg (\figureref{fig:ablations}, right). Whereas with the reward shaping used in our method, with a penalty for being on the ground, agents only learned to get up and stand still. They never learned to walk around or score, despite the inclusion of shaping reward terms for walking forward and towards the ball. These results suggest that in this setting, the exploration problem was too difficult when the agent needed to learn to both get up and play soccer. Our method overcomes this by first separately training policies for these two skills.

\subsubsection*{Importance of Self-Play}
We also ablated the use of self-play in the second stage, while keeping the skill policy regularization and the shaped reward the same. For evaluation, we played agents against a fixed set of six diverse opponents, trained with a variety of approaches; this set includes the final 1v1 agent trained with our full pipeline. \figureref{fig:ablations} shows a comparison of our full training method against two alternatives: training directly against the fixed set of opponents throughout learning, rather than using self-play, and using self-play but sampling opponents from all previous snapshots of the policy, rather than the first quarter. The latter led to unstable learning and converged to poor performance. The former, interestingly, performed slightly worse than agents trained with our self-play method, despite having the advantage of training directly against the opponents used for evaluation.
\section*{}

% Include acknowledgments of funding, any patents pending, where raw data for the paper are deposited, etc.
{\bf Acknowledgments:} We would like to thank Daniel Hennes at Google DeepMind for developing the plotting tools used for the soccer matches, and Martin Riedmiller and Michael Neunert at Google DeepMind for their helpful comments. {\bf Funding:} This research was funded by Google DeepMind. 
{\bf Author contributions:}
Algorithm development: T.H., B.M., G.L., S.H.H., D.T., J.H., M.W., S.T., N.Y.S., R.H., M.B., K.H., A.B., L.H., Y.T., F.S.; 
Software infrastructure and environment development: T.H., B.M., G.L., S.H.H., J.H., S.T., N.Y.S., R.H., M.B., Y.T.; 
Agent analysis: T.H., B.M., G.L., S.H.H.; 
Experimentation: T.H., B.M., G.L., S.H.H., D.T., J.H., N.Y.S.; 
Article writing: T.H., B.M., G.L., S.H.H., D.T., M.W., N.He.; 
Infrastructure support: N.B., F.C., S.S., C.G., N.S., K.P., M.G.;
Management support: N.B., F.C., A.H., N.Hu.;
Project supervision: F.N., R.H., N.He.; 
Project design: T.H., N.He.
{\bf Data availability:} The data used for our quantitative figures and tables has been made available for download at \url{https://zenodo.org/records/10793725}~\cite{zenodo2024}.

\bibliography{SR_bib}
\bibliographystyle{ieeetr}

\appendix
\clearpage
\begin{center}\huge
Supplementary Materials
\end{center}
% {\let\newpage\relax\maketitle}

\subsection*{Contents}

\beginsupplement
\begin{tabular}{p{0.23\textwidth}p{0.68\textwidth}}
    Suppl. Methods &  \\
    \xfigref{fig:op3} & The OP3 Robot \\
    \xfigref{fig:get_up_key_frames} & Get-Up Skill Training Poses  \\
    \xfigref{fig:learning_curves} & Learning Curves  \\
    \xfigref{fig:reward_ablations} & Reward Ablations  \\
    \xfigref{fig:learned_gait} & Learned Gait  \\
    \xfigref{fig:scripted_gait} & Scripted Gait  \\
    \xfigref{fig:vision_analysis} & Vision Analysis  \\
    \tabref{table:joint_limits} & Joint Limits \\
    \tabref{table:observations} & Observations \\
    \tabref{table:ShapingRewards} & Reward Components \\
    \tabref{table:hyperparams} & Hyperparameters \\
    \tabref{table:training_time} & Training Time \\
    \supmovref{\movtraininginsim} & Episodes from training in simulation of the skills and 1v1 policy.\\
    \supmovref{\movmatches} & Compilation of several continuous real-world 1v1 matches.\\
    \supmovref{\movsetpieces} & Compilation of penalty kick and get-up-and-shoot set pieces.\\
    \supmovref{\movpushes} & Examples of recovery from pushes and falling.\\
    \supmovref{\movvision} & Description of training and set pieces for vision-based agents.\\
\end{tabular}

\clearpage

\section*{Supplementary Methods}
\label{app:supplementary_methods}

\subsection*{Related Work}
\label{app:extended_related_work}

\subsubsection*{Skill and Transfer Learning}
Skill learning and transfer of learned behaviors has been a long-standing active area of research \citep{thrun1994finding,bowling1998reusing,stone2000layered,macalpine2018overlapping}. The data requirements for training neural networks as policies have recently emphasized the requirement for increased data efficiency. Different mechanisms for transfer have been proposed, ranging from direct reuse of parameters in flat or hierarchical agents \citep{peng2019mcp,wulfmeier2021data,parisotto2015actor,WonControlStrategies,sutton1999between,salter2022mo2}, auxiliary objectives \citep{ross2011reduction, Tirumala2020priorsjournal, LiuHumanoidSoccer}, to transfer via a skill's generated experience \citep{Riedmiller2018sacx,vezzani2022skills}.
A related line of work is kickstarting, which makes use of a trained teacher policy to enable a student policy to learn more quickly and obtain better performance, on the same task \citep{Schmitt2018KickstartingDR,Galashov2019infoasym,abdolmaleki2021dime, Team2023HumanTimescaleAI}. These approaches transfer knowledge from the teacher to the student via a distillation loss, defined as either the cross-entropy or KL-divergence between the output of the student and teacher networks \citep{parisotto2015actor,rusu2015policy}. \citet{abdolmaleki2021dime} frames kickstarting as a multi-objective problem, that must trade off between regularization to the teacher policy versus the RL objective of maximizing expected return. Our approach builds on the linear scalarization (or weighted-sum) approach to combining reward and kickstarting proposed in \citet{abdolmaleki2021dime}, and we follow their proposal to adjust the level of regularization to the teacher depending on the student's performance. We extend this approach to work in our setting, where we have more than one teacher, or skill policy. In particular on real robotics platforms, skill transfer has been critical for increased data efficiency \citep{hafner2021towards, wulfmeier2019compositional}.

\subsubsection*{Multi-agent Reinforcement Learning}
Our approach to training against a mixture of previous opponents is motivated by established methods for multi-agent training. Reinforcement learning with pure self-play can exhibit unstable or cyclic behavior, since learning focuses on the exploitation of a single current policy \citep{BalduzziOpenEnded}, and by overfitting can become exploitable. Therefore many applications of reinforcement learning to multi-agent domains train against a mixture of opponents. In normal form games Fictitious Play \citep{BrownFictitiousPlay}---in which successive best responses to the mixture of previous policies are computed---converges to a Nash equilibrium for two-player, zero-sum games, and it has been generalized to extensive form games using reinforcement learning by, for example, \citet{lanctot2017unified,HeinrichFictitious}. Alternatively, but similarly, \citet{VinyalsStarCraft} achieved stability and robustness by playing against a league of opponents. Our work is an efficient implementation of these ideas since we train against a mixture of previous opponents, but using one continuous training epoch, rather than successive generations. This improves efficiency, but could carry an increased risk of getting stuck in a local optima. Orthogonally, self-play provides a natural auto-curriculum in multi-agent RL \citep{bansal2017emergent, BakerToolUse}, which can be important since finding the best response to a set of strong opponents from scratch can be challenging for RL in some domains. For example, in soccer, strong opponents could dominate play and a learner might get very little experience of ball interaction. Our method effectively features a similar auto-curriculum property, since we trained in one continuous epoch in which the opponents are initially weak, and, subsequently, are automatically calibrated to the strength of the current agent as earlier agent checkpoints are successively added to the opponent pool.

\subsection*{Environment Details}
\label{app:environment_details}
\supfigureref{fig:op3} shows one of the OP3 robots we used in our experiments. The robot has mechanical improvements to reduce damage from testing different, potentially unsafe, policies on the robot, including 3D printed bumpers and arms. We also installed a 3D printed vest for motion capture markers for tracking. We limited the achievable joint range programmatically to reduce the risk of self-collisions, as listed in \suptableref{table:joint_limits}. The limits are applied to each joint independently. The joints corresponding to the left and right side are prefixed with ``l'' and ``r'', respectively. Zero joint angles correspond to the T-pose. Note that these limits don't completely eliminate collisions, as we wanted to keep the limits wide enough to allow agile and dynamics behaviors. \suptableref{table:observations} lists all observations that are available for the agent.

\begin{minipage}{\textwidth}
  \begin{minipage}[b]{0.49\textwidth}
    \begin{figure}[H]
      \centering
      \includegraphics[width=0.65\textwidth]{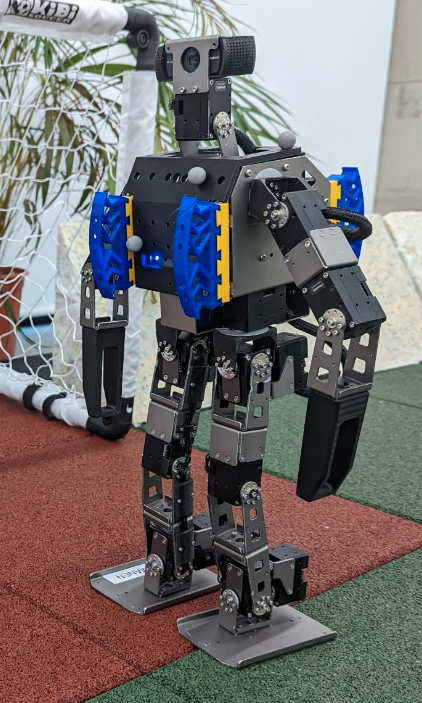}
      \caption{\textbf{OP3 robot.} One of the OP3 robots we used in our experiments.}
      \label{fig:op3}
    \end{figure}
  \end{minipage}
  \hfill
  \begin{minipage}[b]{0.49\textwidth}
    \begin{table}[H]
      \centering
      \fontsize{10}{11}\selectfont
      \begin{tabular}{rlll}
        \toprule \midrule
        \textbf{ID} & \textbf{Joint Name} & \textbf{Min} & \textbf{Max} \\
        \toprule
        19 & head\_pan    & -0.79 &  0.79 \\
        20 & head\_tilt    & -0.63 & -0.16 \\
        16 & l\_ank\_pitch & -0.4  &  1.8  \\
        18 & l\_ank\_roll  & -0.4  &  0.4  \\
        6  & l\_el         & -1.4  &  0.2  \\
        12 & l\_hip\_pitch & -1.6  &  0.5  \\
        10 & l\_hip\_roll  & -0.4  & -0.1  \\
        8  & l\_hip\_yaw   & -0.3  &  0.3  \\
        14 & l\_knee       & -0.2  &  2.2  \\
        2  & l\_sho\_pitch & -2.2  &  2.2  \\
        4  & l\_sho\_roll  & -0.8  &  1.6  \\
        16 & r\_ank\_pitch & -1.8  &  0.4  \\
        17 & r\_ank\_roll  & -0.4  &  0.4  \\
        5  & r\_el         & -0.2  &  1.4  \\
        11 & r\_hip\_pitch & -0.5  &  1.6  \\
        9  & r\_hip\_roll  &  0.1  &  0.4  \\
        7  & r\_hip\_yaw   & -0.3  &  0.3  \\
        13 &r\_knee        & -2.2  &  0.2  \\
        1  &r\_sho\_pitch  & -2.2  &  2.2  \\
        3  &r\_sho\_roll   & -1.6  &  0.8  \\
        \midrule
        \bottomrule
      \end{tabular}
      \caption{Joint Limits}
      \label{table:joint_limits}
    \end{table}
  \end{minipage}
\end{minipage}

\begin{table}[H]
    \centering
    \small
    \begin{tabular}{lllp{9cm}} 
        \toprule
        & \textbf{Observation} & \textbf{Dimension} & \textbf{Note} \\ %
        \toprule
        \parbox[t]{2mm}{\multirow{5}{*}[-0.4em]{\rotatebox[origin=c]{90}{Proprioception}}}
        & joint positions & $5\cdot20$ &Joint positions in radians (stacked last 5 timesteps) \\
        & linear acceleration &$5\cdot3$&Linear acceleration from IMU (stacked)\\
        & angular velocity &$5\cdot 3$  &Angular velocity (roll, pitch, yaw) from IMU (stacked)\\
        & gravity &$5\cdot 3$& Gravity direction, derived from angular velocity and linear acceleration using Madgwick filter (stacked)\\
        & previous action &$5\cdot20$& Action filter state (stacked)\\
        \midrule
        \parbox[t]{2mm}{\multirow{7}{*}{\rotatebox[origin=c]{90}{Game State}}}
        &  ball position &$2$& \multirow{7}{*}{\parbox[t]{9cm}{All positions and velocities are 2-dimensional vectors corresponding to the planar coordinates expressed in the agent's frame. The velocities are derived from positions via finite differentiation. Agent velocity is stacked over five timesteps.}}\\
        &  opponent position&$2$&\\
        &  goal position&$2$&\\
        &  opponent goal position&$2$&\\
        &  agent velocity &$5\cdot2$&\\
        &  ball velocity &$2$&\\
        &  opponent velocity &$2$&\\
    \end{tabular}
    \caption{The agent's observations.}
    \label{table:observations}
\end{table}

\subsection*{Training Details}
\label{app:training_details}

\subsubsection*{Reward Functions}
\label{app:rewards}
Reward components for training the skills and the final policy is given in \suptableref{table:ShapingRewards}. The last three columns denote the weight of each reward component constituting the total reward for each training stage.

\begin{table}[H]
    \centering
    \small
    \begin{tabular}{llp{8.3cm}ccc}
        \toprule
        \textbf{Type} & \textbf{Reward $\hat{r}_k$} & \textbf{Description and purpose} &  %
        \rotatebox[origin=c]{90}{\hspace{0.1em}Soccer skill} &
        \rotatebox[origin=c]{90}{\hspace{0.1em}Get-up skill} &
        \rotatebox[origin=c]{90}{\hspace{0.3em}Full 1v1 agent}\\
        \toprule
        \vspace{0.2cm}
        \parbox[t]{2mm}{\multirow{5}{*}[-5.7em]{\rotatebox[origin=c]{90}{Shaping}}}
        & Scoring & $+1$ when the agent scores a goal and $0$ otherwise. & 1000 & --- & 1000\\% & \ref{XXX} \\
        \vspace{0.2cm}
        & Conceding & $-1$ when the opponent scores a goal and $0$ otherwise. & 0 & --- & 1000 \\% & \ref{XXX} \\
        \vspace{0.2cm}
        & Velocity to ball & The magnitude of the agent's velocity toward the ball. & 0.05 & --- & 0.05 \\ %
        \vspace{0.2cm}
        & Velocity & The magnitude of the player's forward velocity. & 0.1 & --- & 0.1 \\ %
        \vspace{0.2cm}
        & Interference & A penalty, equal to the cosine of the angle between the player's velocity vector and the heading of the opponent, if the player is within \SI{1}{\meter} of the opponent. This discourages the agents from interfering with and fouling the opponent. & 1 & --- & 1 \\ %
        \vspace{0.2cm}
        & Termination & A penalty, equal to $-1$ if the player is on the ground, out of bounds, or in the goal penalty area. This encourages agents to get up from the ground quickly, avoid falling, and stay in bounds. & --- & --- & 0.5 \\ %
        \midrule
        \vspace{0.2cm}
        \parbox[t]{2mm}{\multirow{2}{*}[-3.1em]{\rotatebox[origin=c]{90}{Sim-to-real}}}
        & Upright & $0$ if the robot is upside down or if the tilt angle is greater than 0.4 radians. Increases linearly, and is equal to $+1$ if the tilt angle is less than 0.2 radians. & 0.015 & --- & 0.02\\ %
        \vspace{0.2cm}
        & Joint torque & A penalty, equal to the magnitude of the torque measured at the player's knees. This discourages the player from learning gaits which cause high forces on the knees, for example during ground impacts, which can damage a physical robot. & 0.01 & --- & 0.01 \\ %
        \midrule
        \vspace{0.2cm}
        \parbox[t]{2mm}{\multirow{1}{*}[-3.3em]{\rotatebox[origin=c]{90}{Skill Training}}}
        \vspace{0.2cm}
        & Target pose & The negative of the product of joint angle error and gravity error, where the reference target pose is sampled from a scripted get-up trajectory. Joint angle error is the L2 norm of the difference of target joint angles and actual joint angles, scaled and clipped to $[0, 1]$, where $0$ denotes an error of $\pi$ and  $1$ denotes no error. Similarly, gravity error is the angle between the target gravity vector and the actual gravity vector, scaled and clipped to $[0, 1]$, where $0$ denotes an error of $\pi/2$ and $1$ denotes full alignment. & --- & 1 & ---\\
        \bottomrule
    \end{tabular}
    \caption{The reward components used for training the skills and the full 1v1 agent.}
    \label{table:ShapingRewards}
\end{table}

\subsubsection*{Get-up Skill Training}
\label{app:getupskill}
\begin{figure}[H]
  \centering
  \includegraphics[width=1.0\textwidth]{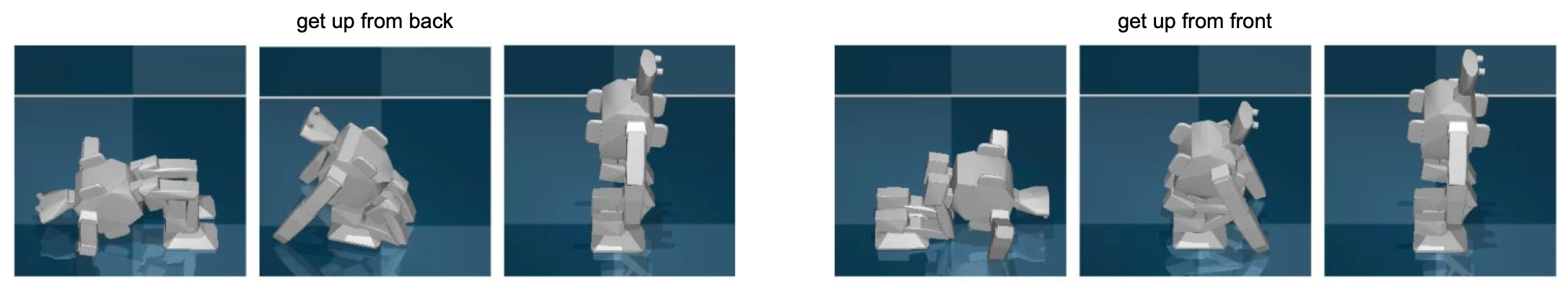}
  \caption{\textbf{Get-up poses.} The key poses used to train the get-up skill, taken from a scripted get-up controller \citep{op3-driver}.}
  \label{fig:get_up_key_frames}
\end{figure}

\supfigureref{fig:get_up_key_frames} shows the three key poses for getting up from the front and back we used to guide training of the get-up skill.

\subsubsection*{Agent Training and Architecture}
\label{sec:Appendix:Training:MPO}
\paragraph{Distributional MPO:}
We trained agents using MPO with a distributional critic, which we refer to as Distributional MPO (DMPO). MPO, or Maximum a Posteriori Policy Optimization \citep{Abdolmaleki_2018}, is an actor-critic deep RL algorithm based on policy iteration. Policy iteration algorithms alternate between performing policy evaluation and policy improvement. Policy evaluation estimates the Q-function (or critic) given the current policy (or actor). Policy improvement updates the policy given the current Q-function.

\paragraph{Policy Evaluation:}
In policy evaluation, the Q-function is learned for the current policy $\pi_\text{t}$, parameterized by the target policy network. Any off-the-shelf Q-learning algorithm can be used to learn the Q-function. We chose to use n-step Q-learning \citep{SuttonBartoBook} with $n=5$. In addition, we parameterize the Q-function as a deep feed-forward neural network that outputs a categorical distribution, as introduced in \citet{Bellamare_2017}.

\paragraph{Policy Improvement:}
In policy improvement, the aim is to improve the policy with respect to the current Q-function $Q_t$ and current policy $\pi_t$. MPO has a two-step policy improvement procedure. The first step computes an improved action distribution $q(\boldsymbol{a}|\boldsymbol{s})$ that optimizes the standard RL objective together with a KL-divergence constraint to the current policy:
\begin{align}
  \max_{q} \; & \mathbb{E}_\xi \bigg[ \int_{\boldsymbol{a}} q(\boldsymbol{a} | \boldsymbol{s} ) Q(\boldsymbol{s},\boldsymbol{a}) \text{d}\boldsymbol{a} \bigg] \label{eq:mpo_policy_improvement_step1} \\ 
  \text{s.t.} \; & \mathbb{E}_\xi \Big[ \text{KL} \big(q(\cdot|\boldsymbol{s}) \| \pi_t (\cdot | \boldsymbol{s}) \big) \Big] < \epsilon \, . \nonumber
\end{align}
The KL-divergence constraint ensures that the improved action distribution does not deviate too much from the current policy, which stabilizes learning. \eqref{eq:mpo_policy_improvement_step1} has a closed-form solution for each state $\boldsymbol{s}$, where
\begin{equation}
    q(\boldsymbol{a} | \boldsymbol{s}) \propto \pi_t(\boldsymbol{a} | \boldsymbol{s}) \exp \bigg( \frac{Q(\boldsymbol{s}, \boldsymbol{a})}{\eta} \bigg) \, ,
\end{equation}
and the temperature $\eta$ is the solution to the convex dual function involving $\epsilon$:
\begin{equation}
    \eta = \argmin_\eta \eta \epsilon + \eta \mathbb{E}_\xi \Bigg[ \log \int_{\boldsymbol{a}} \pi_t(\boldsymbol{a}|\boldsymbol{s}) \exp\bigg(\frac{Q(\boldsymbol{s}, \boldsymbol{a})}{\eta} \bigg) \text{d}\boldsymbol{a} \Bigg] \, .
\end{equation}
In practice, rather than setting $\eta$ to the exact solution of the convex dual function, it is effective to optimize $\eta$ via gradient descent, initializing with the solution found in the previous iteration.

The second step distills the improved action distribution $q$ into a new parametric policy $\pi_{t+1}$ (with parameters $\theta_{t+1}$) via a supervised learning loss:
\begin{align}
    \theta_{t+1} = \argmax_\theta \; & \mathbb{E}_\xi \bigg[ \int_{\boldsymbol{a}} q(\boldsymbol{a}|\boldsymbol{s}) \log \pi_\theta(\boldsymbol{a}|\boldsymbol{s}) \text{d}\boldsymbol{a}  \bigg] \\
    \text{s.t.} \; & \mathbb{E}_\xi \Big[ \text{KL}\big( \pi_t(\cdot|\boldsymbol{s}) \| \pi_\theta(\cdot|\boldsymbol{s}) \big) \Big] < \beta \, . \nonumber
\end{align}
The additional KL-divergence constraint enforces that the policy does not change too much in a single iteration of policy improvement, which avoids premature convergence and improves learning stability \citep{Schulman_2015,Abdolmaleki_2018}.

\paragraph{Agent Hyperparameters:}
\label{sec:Appendix:Training:Architecture}
\suptableref{table:hyperparams} lists the hyperparameters we used for agent architectures and training. Both the policy and critic were feed-forward neural networks. During training we sample a batch of trajectory segments from the replay buffer, of size $B$ and segment length $T$, perform one step each of policy evaluation and policy improvement on this batch, and repeat. To stabilize learning, we maintain a separate online and target network for the policy and the Q-function \citep{Mnih_2015}. The policy evaluation and improvement steps update the parameters of the online network, and the online network parameters are copied to the target network at fixed intervals.

\begin{table}[t]
\centering
\vskip 0pt
\fontsize{10}{11}\selectfont
\setlength\tabcolsep{0pt}
\begin{tabular}{lc} 
 \toprule \midrule
 \textbf{Hyperparameter} & \textbf{Setting}  \\
 \midrule
 Q-function / critic network: & \\
 \hspace{2em} layer sizes & $(400, 400, 400, 300)$ \\
 \hspace{2em} support & $[-150, 150]$ \\
 \hspace{2em} number of atoms & $51$ \\
 \hspace{2em} n-step returns & $5$ \\
 \hspace{2em} discount factor $\gamma$ & $0.99$ \\
 \midrule
 policy / actor network: & \\
 \hspace{2em} layer sizes & $(256, 256, 128)$ \\
 \hspace{2em} minimum variance & $0.001$ \\
 \midrule
 policy and Q-function networks: & \\
 \hspace{2em} layer norm on first layer? & yes \\ 
 \hspace{2em} $\tanh$ on output of layer norm? & yes \\
 \hspace{2em} activation (after each hidden layer) & ELU \\
 \midrule
 MPO for policy improvement: & \\
 \hspace{2em} actions sampled per state & $20$ \\
 \hspace{2em} $\epsilon$ & $0.1$ \\
 \hspace{2em} KL-constraint on mean, $\beta_{\mu}$ & $0.0025$ \\
 \hspace{2em} KL-constraint on covariance, $\beta_{\Sigma}$ & $10^{-6}$ \\
 \hspace{2em} Adam learning rate for dual variables & $10^{-2}$ \\
 \hspace{2em} Adam learning rate for policy and Q-function networks & $10^{-4}$ \\
 \midrule
 training & \\
 \hspace{2em} batch size, $B$ & $256$ \\ 
 \hspace{2em} trajectory segment length, $T$ & $48$ \\ 
 \hspace{2em} replay buffer size & $10^{6}$ \\
 \hspace{2em} target network update period for actor & $25$ \\ 
 \hspace{2em} target network update period for critic & $100$ \\ 
 \midrule
 \bottomrule
\end{tabular}
\caption{Hyperparameters for agent architectures and training.}
\label{table:hyperparams}
\end{table}

\subsubsection*{Training Curves}
\label{app:training_curves}
\begin{figure}
  \centering
  \includegraphics[width=\textwidth]{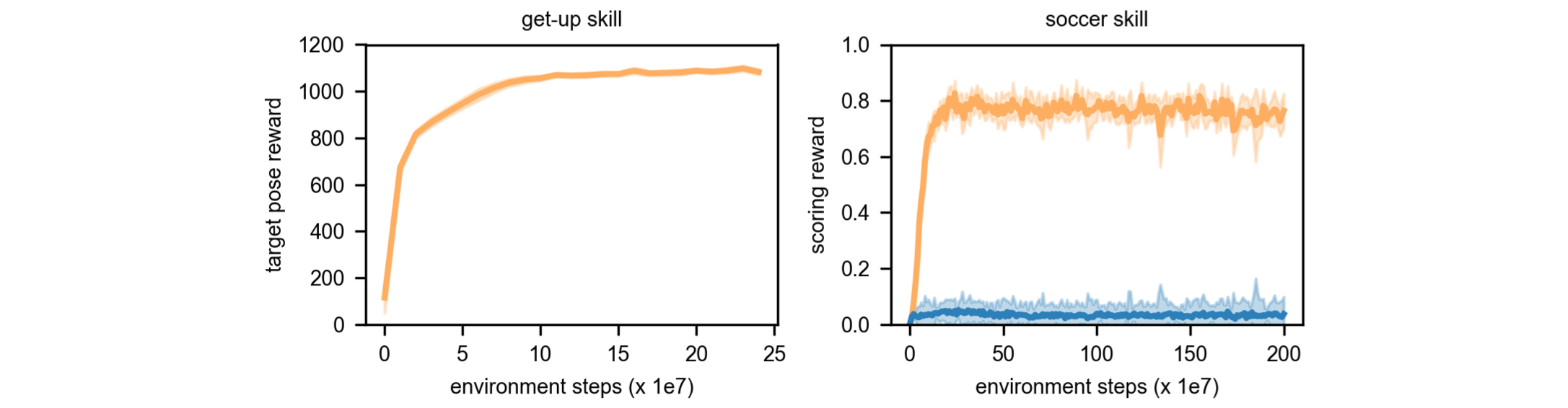}
  \caption{\textbf{Learning curves for the get-up and soccer skills.} (\figureref{fig:ablations} shows learning curves for the full 1v1 task.)  The get-up and soccer policies at the end of training are used as the pre-trained skills for the 1v1 task. For the soccer agent, the orange line denotes the scoring reward of the agent, where 1 corresponds to scoring a goal. The blue line denotes the negative of the conceding penalty, where 1 corresponds to the opponent scoring a goal. The error bars show 95\% confidence intervals over ten seeds.}
  \label{fig:learning_curves}
\end{figure}

\supfigureref{fig:learning_curves} shows learning curves for training the skill policies in simulation, and \suptableref{table:training_time} shows the time spent training each agent. Both the get-up and soccer skills learned relatively quickly. Although the learning curve plateaus, we found that soccer agents trained for longer were better at anticipating ball movement and handling the ball, for instance kicking a moving ball.

The control frequency is 40 Hz, so each environment step corresponds to $0.025$ seconds of simulation time. Wall-clock time refers to the amount of time it takes to train the agent. This is significantly shorter than the simulation time because we use distributed training, in that multiple copies of the agent interact with the simulated environment in parallel.

\begin{table}[t]
    \centering
    \small
    \begin{tabular}{l@{\hspace{2\tabcolsep}} c@{\hspace{1.5\tabcolsep}}c@{\hspace{1.5\tabcolsep}}c}
    \toprule
    Agent & Environment Steps & Simulation Time & Wall-Clock Time \\
    \midrule
    Get-up Skill & $2.4 * 10^8$ & $70$ days & $14$ hours \\
    Soccer Skill & $2.0 * 10^9$ & $580$ days & $158$ hours \\
    Full 1v1 Agent & $9.0 * 10^8$ & $262$ days & $68$ hours \\
    \bottomrule
    \end{tabular}
    \caption{The time spent training each agent.}
    \label{table:training_time}
\end{table}

\begin{figure}
  \centering
  \includegraphics[width=\textwidth]{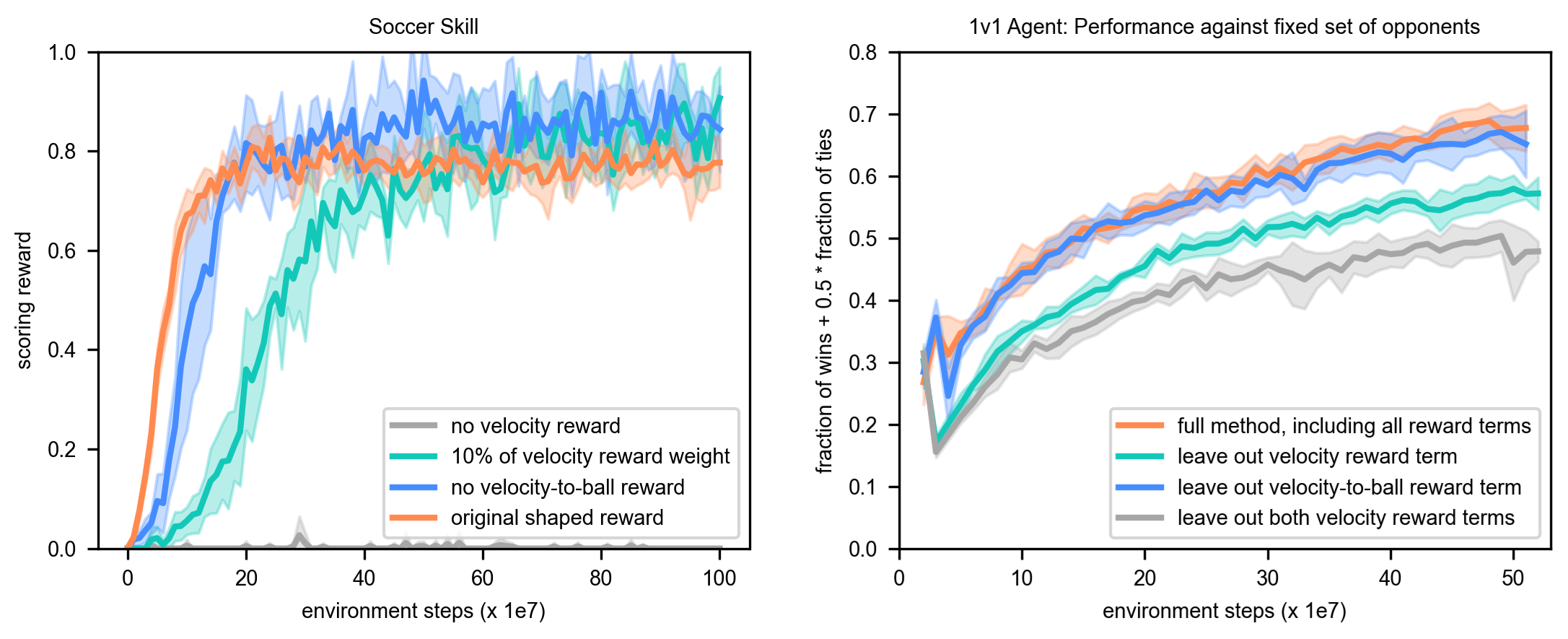}
  \caption{\textbf{Reward term ablations.} Ablations of the velocity-based reward terms for training the soccer skill (left) and the full 1v1 agent (right).}
  \label{fig:reward_ablations}
\end{figure}

\subsubsection*{Reward Ablations}
\label{app:reward_ablations}
The reward we use for training the skills and full 1v1 agent consists of several components. We ran ablations to investigate which components are necessary for learning. \supfigureref{fig:reward_ablations} shows the learning curves for the soccer skill and full 1v1 agent, when the forward velocity reward component and/or the velocity-to-ball component is ablated. Recall that the soccer skill is trained against an untrained opponent, so scoring is much easier. We chose to ablate only the velocity-based reward components because the other reward components are either necessary for sim-to-real transfer or for defining the task (for example, scoring and staying in-bounds). For both the soccer skill and full 1v1 agent, leaving out the forward velocity reward hinders learning; in fact the soccer skill does not learn at all. This suggests that the forward velocity component is important for making exploration easier. For the soccer skill, leaving out the velocity-to-ball reward or using a smaller weight on the forward velocity reward delays learning initially, but the policies eventually converge to slightly higher performance. For the full 1v1 agent though, leaving out the velocity-to-ball reward converges to slightly worse performance.

\subsection*{Behavior Analysis Experiment Details}
\label{app:behavior_analysis_details}

\subsubsection*{Baseline Behavior Comparisons: Experiment Details}
\label{app:baseline_comparison_details}

Here we provide experiment details for the results in \nameref{sec:baseline_comparison}.

\paragraph{Walking:} To compare walking speed, we ran 10 episodes with both the learned agent and the scripted walking controller. For each episode, we initialized the robot in the standard ``walking ready'' pose, and ran both the agent or baseline for 10 seconds. We measured the speed using the distance traveled over the five second window of time starting from the 2nd second to the 7th second. Since the scripted controller does not actively correct the robot's heading, it often quickly veered off course. Therefore, to make the comparison fair for the baseline, we approximated the distance traveled over the five second window by summing the distances traveled in each of the 5 one second windows comprising the whole window.

The agent does not have a specific walking behavior that can be executed in isolation, so we encouraged it to walk forwards by initializing the robot close to its own goal, and placed the ball in the opponent's goal area.

\paragraph{Get-Up Ability:} 
Similar to walking, the learned agent does not have a specific get-up behavior that can be executed in isolation. However, the agent prefers to stay upright, so simply initializing the robot on the floor triggers the get-up behavior. To compare the get-up behaviors, we thus placed the robot on the floor, face down in a T-pose (where all joint positions are set to zero), and activated either the agent or the scripted get-up controller. We considered the robot as standing as soon as it reached a height of \SI{36}{\cm}, measured at the motion capture marker on the shoulder. Note that when the robot is fully upright, the motion capture marker on the shoulder reaches a height of \SI{41}{\cm}. Hence a threshold of \SI{36}{\cm} corresponds to approximately a \SI{5}{\cm} tolerance in shoulder height. This is necessary since in general the learned policy, after getting up, quickly starts moving away from the standing position, towards the ball for example. We collected 10 episodes from both behaviors and compared the time required to reach the specified get-up threshold. Both the scripted behavior and the agent succeeded in standing up on every trial.

\paragraph{Kicking Speed:} To compare kicking ability, we placed the ball \SI{1}{\m} from the opponent's goal and placed the robot behind the ball, facing the goal. For the scripted controller, which is ``blind'' in the sense that it does not take into account the ball's location, we were careful to place the robot such that the swung leg was exactly behind the ball, to maximize the impact. The learned policy is conditioned on the ball position, so we aligned the ball between the legs, allowing the agent to choose which leg to use for the kick. We collected 10 episodes from both behaviors. The learned policy attempted to kick the ball directly towards the goal on every trial. Initializations in which some other strategy (for instance dribbling) would be chosen are easily avoided. We determined the kicking speed by calculating the total distance traveled in the \SI{0.2}{\s} window following the first contact with the ball. We observed that the learned policy can kick more powerfully if we initialize the ball further away from the robot, to allow it to take a few steps before the kick. Therefore, we collected 10 additional episodes with the ball \SI{0.5}{\m} from the goal and an approach distance of \SI{2.5}{\m}. This latter setup was not possible to replicate with the scripted controller, since it cannot take into account the ball's location.

\paragraph{Turning Speed:} To compare turning ability, we placed the ball \SI{1}{\m} from the opponent's goal and placed the robot in its own half \SI{1.5}{\m} from its own goal, facing 180 degrees away from the ball. We determined the turning speed by measuring the time taken to change orientation from $\pm$45\degree to $\pm$135\degree (depending on the direction the agent turned). For the scripted controller distributed with the OP3, we optimized the step length and step speed parameters to maximize turning speed while remaining upright on most trials. Episodes in which the robot fell were discounted and re-run, until we had collected 10 episodes from both behaviors. The learned behavior fell in 3 out of 13 trials, whereas the scripted controller remained upright on 10 consecutive trials.

\subsubsection*{Set Piece: Experiment Details}
\label{app:set_piece_details}

Here we provide additional experiment details for the results in \nameref{sec:behavior_analysis}.

The get-up-and-shoot set piece is a short episode of 1v1 soccer in which the ball is initialized at the center of the pitch (\SI{2.5}{\m} from the goal), and the agent is initialized face down in a default T-pose behind the ball at a random position. The initial position of the robot was sampled uniformly from a grid of positions in the center of the robot's own half; the positions used in the simulation and real environments were identical. The opponent is initialized in a T-pose \SI{0.5}{\m} away from the goal line and the sideline, in its own half, and remains stationary throughout the episode. The episode ends after \SI{10} seconds or after a goal is scored. Example initial configuration is shown in \figureref{fig:set_piece_analysis}.

\subsection*{Learned Gaits}
\label{app:learned_gait}
\begin{figure}[H]
  \centering
  \includegraphics[width=1.0\textwidth]{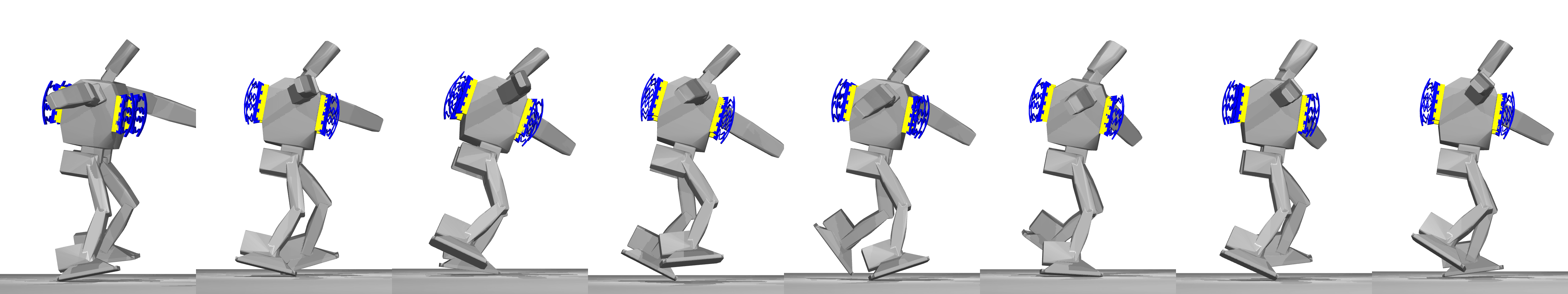}
  \caption{\textbf{Policy walking.} Stills 100ms apart, walking with the Deep RL policy.}
  \label{fig:learned_gait}
\end{figure}

\begin{figure}[H]
  \centering
  \includegraphics[width=1.0\textwidth]{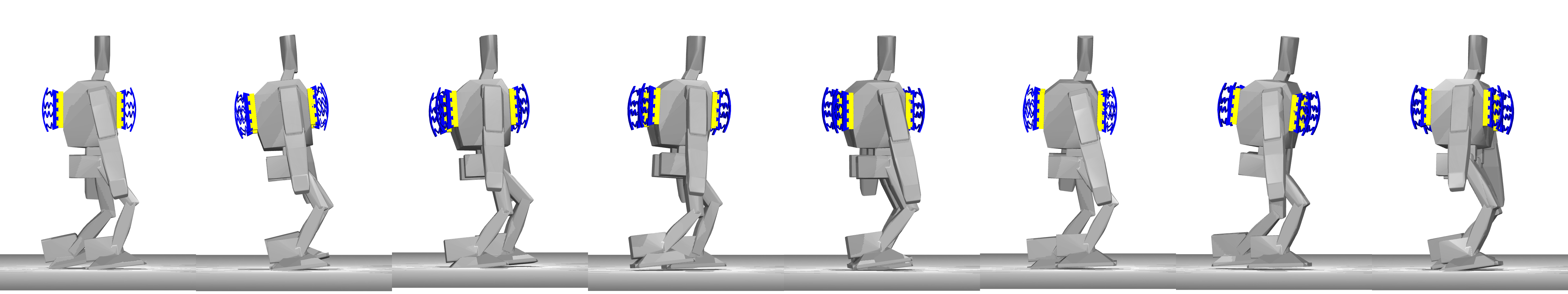}
  \caption{\textbf{Baseline controller walking.} Stills at 96ms apart, straight line walking with the controller of \citep{op3-driver}.}
  \label{fig:scripted_gait}
\end{figure}

\supfigureref{fig:learned_gait} and \supfigureref{fig:scripted_gait} show representative sequences of walking gaits obtained from the Deep RL learned policy, and from the scripted walking controller of \citep{op3-driver} respectively.  The scripted controller behaves conservatively and remains close to statically stable configurations.  By contrast, the learned gait makes more aggressive use of the robot dynamics by using the arms to balance and making contact primarily with the edges of the feet.

\subsection*{Playing Soccer from Raw Vision}
\label{app:vision}

\subsubsection*{Training Details}
To create a realistic visual simulation of the real world, we created a NeRF model \citep{mildenhall2021nerf} of our soccer pitch inspired by the approach of \citet{NeRF2Real}. The NeRF model captures static properties of the scene including the background, goal locations and the pitch itself. We overlaid these images with renders from the MuJoCo camera of dynamic objects, that is, the ball and opponent. For robustness, we randomized over visual properties of the ball, including color. In addition, we randomized training over NeRF models captured at different points in time to be robust to changes in the background and lighting conditions. The reward and environment setup was otherwise identical to that used for state. To increase learning efficiency, for every iteration of training a new agent, we mixed offline data gathered from previous iterations with online replay data. 

We used the same agent (MPO with distributional critic) for vision as for our state-based analysis. The policy was conditioned on visual information and proprioception while the critic was given the same information as in state-based training. For our policy we used the same network configuration as \citet{NeRF2Real}.

\subsubsection*{Set Piece Analysis}
\begin{figure}
  \centering
  \includegraphics[width=0.7\textwidth]{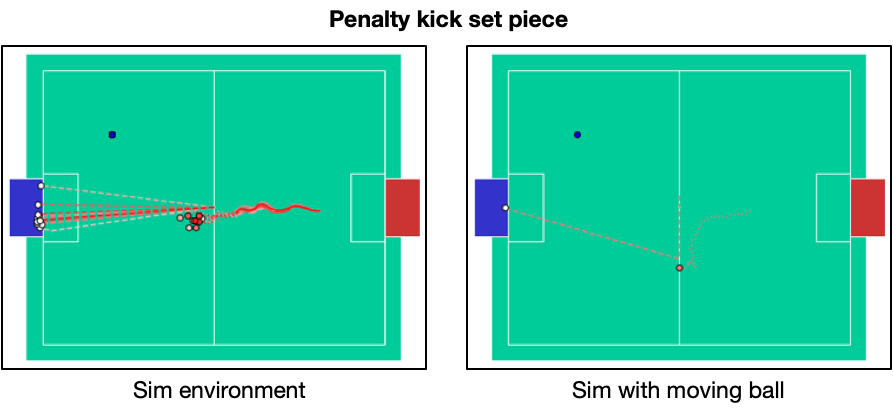}
  \caption{\textbf{Set piece with the vision-based policy.} Overlaid plots of the trajectories collected from the simulated vision set piece experiment. We refer the reader to our supplementary website for videos of experiments on real robots. Robot trajectory before kick (solid red lines), after kick (dotted red line) and corresponding ball trajectory (dashed red line), ball (white circle) and opponent (blue circle). (Left) Analysis of penalty kick from fixed starting positions of ball and robots. The robot scored in all trials. (Right) A single trajectory showing adaptive movement behavior when the ball is initialized to be rolling along the negative y-axis.}
  \label{fig:vision_analysis}
\end{figure}

We conducted a simulated penalty shootout using a vision trained policy as shown in \supfigureref{fig:vision_analysis} (left).  The penalty kick set piece is a short episode of 1v1 soccer in which the ball is initialized at the center of the pitch, offset \SI{1.5}{\m} behind the ball closer to its own goal. The opponent is initialized \SI{1}{\m} away from the goal line and the sideline, in its own half, and remains stationary throughout the episode. The episode ends after 6 seconds or after a goal is scored. The robot consistently scored across 10 trials in the set piece configuration. The robot scored 6 out of 10 times (\SI{60}{\percent}) when transferred to the real world and hit the post 3 times. Variations in the visual elements of the scene and sensor noise could explain the discrepancy. 

In addition in \supfigureref{fig:vision_analysis} (right) we plot a trajectory where the ball is initialized moving in the negative y-direction. The robot predicts the movement of the ball and proactively alters its gait to strike at a particular location. Examples of agent behavior from vision can be found in \supmovref{\movvision}.

\end{document}